\documentclass[sn-mathphys-num]{sn-jnl}

\usepackage[dvipsnames]{xcolor}%
\usepackage{adjustbox}%
\usepackage{algorithm,algorithmicx}%
\usepackage{algpseudocode}%
\usepackage{amsmath,amssymb,amsfonts}%
\usepackage{amsthm}%
\usepackage{graphicx}%
\usepackage{multirow}%
\usepackage{mathrsfs}%
\usepackage[title]{appendix}%
\usepackage{textcomp}%
\usepackage{wrapfig}%
\usepackage{graphicx}%
\usepackage{manyfoot}%
\usepackage{booktabs}%
\usepackage{listings}%
\usepackage{pifont}%
\usepackage{enumitem}%
\usepackage{makecell}%
\usepackage{caption}%
\usepackage{subcaption}%
\usepackage{svg}%
\usepackage{orcidlink}%

\newcommand{\cmark}{\color{ForestGreen}\ding{51}}%
\newcommand{\xmark}{\color{BrickRed}\ding{55}}%

\theoremstyle{thmstyleone}%
\theoremstyle{thmstyletwo}%

\theoremstyle{thmstylethree}%
\newtheorem{definition}{Definition}%

\begin{document}

\title[Article Title]{Transformers  Utilization in Chart Understanding: A Review of Recent Advances \& Future Trends}


\author*[1]{\fnm{Mirna} \sur{Al-Shetairy} \orcidlink{0009-0007-0380-8784}}\email{mshetairy@cis.asu.edu.eg}

\author[1]{\fnm{Hanan} \sur{Hindy} \orcidlink{0000-0002-5195-8193}}\email{hanan.hindy@cis.asu.edu.eg}

\author[2]{\fnm{Dina} \sur{Khattab} \orcidlink{0000-0003-1201-1789}}\email{dina.khattab@cis.asu.edu.eg}

\author[1]{\fnm{Mostafa M.} \sur{Aref} \orcidlink{0000-0002-1278-0070}}\email{mostafa.aref@cis.asu.edu.eg}

\affil[1]{\orgdiv{Computer Science Department}, \orgname{Faculty of Computer and Information Sciences, Ain Shams University}, \orgaddress{\city{Cairo}, \country{Egypt}}}

\affil[2]{\orgdiv{Scientific Computing Department}, \orgname{Faculty of Computer and Information Sciences, Ain Shams University}, \orgaddress{\city{Cairo}, \country{Egypt}}}


\abstract{In recent years, interest in vision-language tasks has grown, especially those involving chart interactions. These tasks are inherently multimodal, requiring models to process chart images, accompanying text, underlying data tables, and often user queries. Traditionally, Chart Understanding (CU) relied on heuristics and rule-based systems. However, recent advancements that have integrated transformer architectures significantly improved performance.
This paper reviews prominent research in CU, focusing on State-of-The-Art (SoTA) frameworks that employ transformers within End-to-End (E2E) solutions. Relevant benchmarking datasets and evaluation techniques are analyzed. Additionally, this article identifies key challenges and outlines promising future directions for advancing CU solutions.
Following the PRISMA guidelines, a comprehensive literature search has been conducted across Google Scholar, focusing on publications from January 2020 to June 2024. After rigorous screening and quality assessment, 32 studies have been selected for in-depth analysis.
The CU tasks are categorized into a three-layered paradigm based on the cognitive task required. Recent advancements in the frameworks addressing various CU tasks are also reviewed. These frameworks are categorized into single-task or multi-task based on the number of tasks solvable by the E2E solution. Within multi-task frameworks, pre-trained and prompt-engineering-based techniques are explored.
Furthermore, this review provides an overview of leading architectures, datasets, and pre-training tasks. Despite significant progress, challenges remain in OCR dependency, handling low-resolution images, and enhancing visual reasoning. Future directions include addressing these challenges, developing robust benchmarks, and optimizing model efficiency. Additionally, integrating explainable AI techniques and exploring the balance between real and synthetic data are crucial for advancing CU research.
}




\maketitle
\section{Introduction}\label{sec1}

Visual representations are powerful tools for conveying information in a compact and understandable way. Data visualization as a discipline encompasses a wide variety of tools used in conveying ideas. Any visual figure could be categorized into either an image, a diagram or a chart ~\citep{huang2024detection}. Charts are a well-known visualization tool across different domains. A chart image is defined as a visual representation of data that helps to summarize its underlying data, reflect insights, and draw conclusions. However, chart images on their own can be ambiguous to the reader and therefore require additional context to be able to extract meaningful information from them. Furthermore, the predominant storage format for chart images on the web is the bitmap format, which poses accessibility challenges for visually impaired users~\citep{shahira2021towards, singh2023towards} The issue with the raw bitmap format is that screen readers cannot parse an image if it lacks an alternative text (alt-text) description, which was reported as a common case according to a report made by WebAIM\footnote{\url{https://webaim.org/projects/million/\#alttext} (Accessed on 22/04/2024)} with 21.6\% of web page’s images missing alt-text. 

\begin{figure}[ht]
 \subfloat[]{
	\begin{minipage}[c][1\width]{
	  0.3\textwidth}
	  \centering
	  \includegraphics[width=1\textwidth]{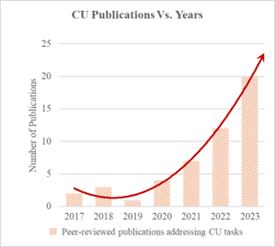}\label{figure-ch1-cu-trends}
	\end{minipage}}
 \hfill 	
 \subfloat[]{
	\begin{minipage}[c][1\width]{
	  0.3\textwidth}
	  \centering
	  \includegraphics[width=1\textwidth]{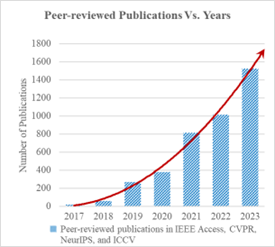}\label{figure-ch1-transformer-trends}
	\end{minipage}}
 \hfill	
 \subfloat[]{
	\begin{minipage}[c][1\width]{
	  0.3\textwidth}
	  \centering
	  \includegraphics[width=1.1\textwidth]{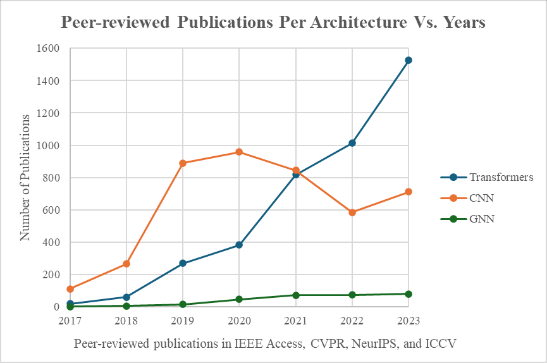}\label{figure-ch1-dl-trends}
	\end{minipage}}
\caption{Statistics on recent research trends obtained through Scopus. (a) Peer-reviewed articles addressing various CU tasks such as CQA, Chart Summarization, and Reasoning over Charts. (b) Frequency of different keywords (e.g., Transformers, BERT and Self-attention) appearing in peer-reviewed articles over the past years in computer science. (c) Articles utilization of different deep learning architectures, namely Transformers, CNNs and GNNs.}\label{fig-ch1-statistics}
\end{figure}

\begin{figure}[h]
\centering
\includegraphics[width=1\textwidth]{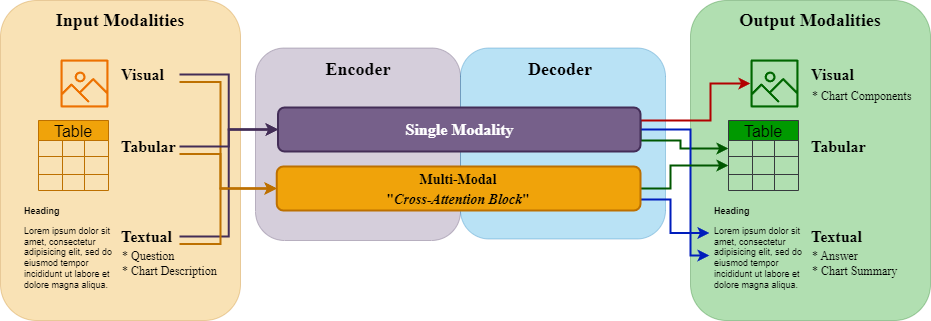}
\caption{An overview of the different modalities in the CU domain. Transformer architectures could either work on a single modality or across multiple ones as input. Each transformer block could then address one of the possible output modalities based on the reviewed body of literature, i.e., addressing different output modalities through a combination of different transformer blocks.}\label{fig-ch1-diff-modalities}
\end{figure}

With the rise of visual language processing, the automation of several chart-focused tasks has witnessed an increased interest, which recently appeared in the literature as Chart Understanding (CU)~\citep{farahani2023automatic, huang2024detection} (refer to Figure~\ref{figure-ch1-cu-trends}). These tasks include but are not limited to Chart Question Answering (CQA), Chart Captioning, and Chart-to-Text Summarization. CU tasks often involve extracting the underlying data from a given chart, such as de-rendering the image into code or its data table. It is considered a subdomain of Visual Language Processing that combines different modalities to retrieve information and build knowledge from chart images (as shown in Figure~\ref{fig-ch1-diff-modalities}). Automating a CU task could involve the usage of a chart image, its data table and/or a user prompt for said task. Similarly, the target of the task could result in any of the previously mentioned modalities. For example, one could potentially interact with a line chart in order to gain a better understanding of its insights. In such cases, the system can result in a modified chart image highlighting the abrupt changes along with a text summarizing the overall line trends. 

\begin{figure}[h]
\centering
\includegraphics[width=0.5\textwidth]{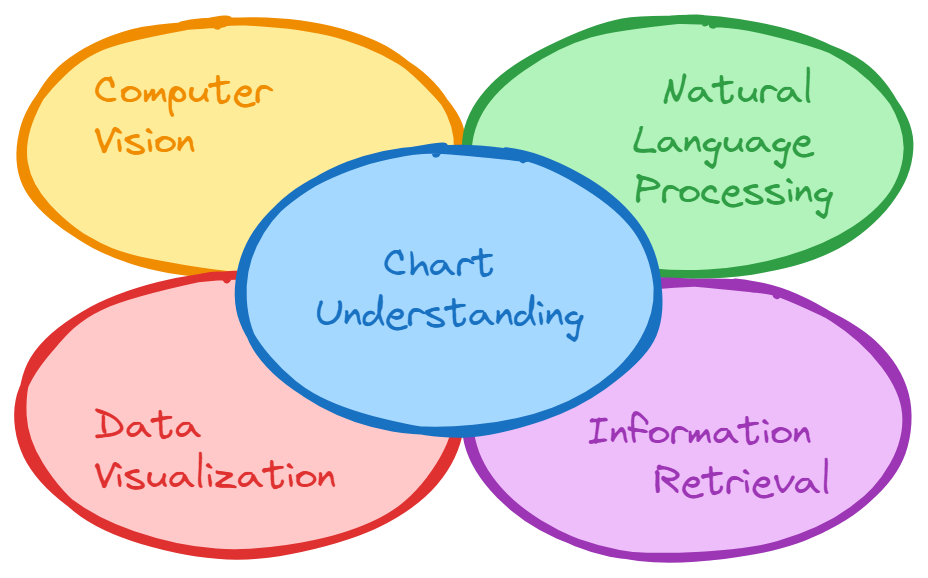}
\caption{Multidisciplinary Nature of the Chart Understanding Domain.}\label{fig-ch1-multidisciplinary-cu}
\end{figure}

Throughout the literature, the emerging discipline of CU could be seen at the heart of four main disciplines, namely, data visualization, Computer Vision (CV), Natural Language Processing (NLP) and information retrieval (refer to Figure~\ref{fig-ch1-multidisciplinary-cu}). A prime example of this intersection is the CQA task. In CQA, a user poses a question (NLP) about a visual element (data visualization and CV) within a chart. To provide a comprehensive answer, the system must not only accurately interpret the visual representation but also access and leverage relevant knowledge bases. The connections between these different disciplines highlight the complexity of CU. To understand CU fully, we need to use knowledge from all of these areas together.

Over the years, numerous applications addressing CU have been developed utilizing different techniques and approaches ranging from using heuristics and rule-based systems to deep learning techniques~\citep{farahani2023automatic}. One emerging approach, in particular, Transformer models~\citep{vaswani2017attention}, is witnessing great development across NLP and CV with an increasing trend of adopting them in vision and multi-modal tasks (as shown in Figure~\ref{figure-ch1-transformer-trends}). This growth motivated us to focus on the applications and utilization in the CU domain. \textbf{Why focus on the transformer architecture?} That would be due to the growing popularity in their utilization within highly regarded venues such as IEEE Access, CVPR, NeurIPS, and ICCV (as shown in Figure~\ref{figure-ch1-dl-trends}). On the other hand, while Convolutional Neural Networks (CNNs) are still being utilized, a decline in usage is observed. Similarly, Graph Neural Networks (GNNs) have limited adoption compared to transformers and CNNs. 

Additionally, we observed the growth of the CU domain across peer-reviewed publications in recent years with a total of 43 articles published within the period of 2020 to 2023 (refer to Figure~\ref{figure-ch1-cu-trends}). Retrieving these preliminary statistics for the CU domain was achieved by looking for the following list of keywords: "Chart Understanding" OR "Chart Question Answering" OR "Chart Summarization" OR "Chart Captioning" OR "Chart Reasoning" OR "Chart To Code" OR "Chart To Table" OR "Chart Data Extraction" OR "Chart Comprehension". In light of these findings, we aim to provide researchers having an interest in multimodal processing and - visual-language domains with a comprehensive review of the transformer's utilization and their recent advancements in the CU domain.

\subsection{Related Surveys}\label{subsec1-1}
This section provides an overview of the latest review articles focusing on chart understanding techniques. A taxonomy of seven data visualization techniques is introduced by~\citep{shakeel2022comprehensive} along with platforms and tools that help produce those visualizations. Building upon existing research in data visualization and automated comprehension, numerous surveys have explored various aspects of visualization identification and understanding tasks. These surveys delve into tasks such as chart recognition and classification~\citep{dhote2023survey}, data extraction~\citep{shahira2021towards}, reconstruction or generation of stylized charts~\citep{ye2024generative}, and question answering or caption generation~\citep{hoque2022chart, bajic2023review, singh2023towards, huang2024detection}. The works of~\citep{wang2021survey, wu2021ai4vis, voigt2022and, shen2022towards, zeng2023review} review advancements of natural language interfaces focusing on visualization generation and recommendation.

Several recent surveys have explored various aspects of CU following the work of~\citep{farahani2023automatic} in which five different tasks were defined as chart detection, chart classification, chart data extraction, chart summarization and chart question answering. These surveys delve into areas like dataset creation and benchmarking~\citep{chen2023state}, human-computer interaction in visualization~\citep{he2024leveraging} and the utilization of foundation models~\citep{yang2024foundation}. In this review, we build upon this foundation by expanding the definitions of chart question answering and summarization based on the insights from these surveys. Additionally, we redefine the chart data extraction task and introduce a new task "chart derendering" to better reflect the broader process of converting the visual representation into other structured formats.

The recent work by~\citep{huang2024detection} presents the closest examination to ours regarding the definition of CU tasks. They offer a valuable overview of techniques used to address CU tasks, categorized by their approach: classification-based, generation-based, or utilizing pre-existing tools. Our work, however, takes a distinct approach. We focus specifically on the development of frameworks that exclusively rely on the transformer architecture throughout their pipeline. We explore the intricate modifications made to these frameworks to effectively adapt them for various CU tasks. This focus on transformer-based adaptations differentiates our work from the broader perspective offered by~\cite{huang2024detection}. We also consider more recent publications and propose an expanded CU taxonomy with three distinct layers encompassing a total of eight CU tasks, including \textit{chart element detection}, \textit{chart data extraction}, \textit{chart type classification}, \textit{chart text role classification}, \textit{quantity alignment across different modalities}, \textit{chart derendering}, \textit{chart question answering}, and \textit{chart-to-text} (refer to Figure~\ref{fig-ch3-taxonomy}).

\subsection{Survey Methodology}\label{subsec1-2}
A systematic search strategy has been followed while building up this research regarding the identification of the relevant literature. A preliminary search through Google Scholar was made using the keywords: “transformer”, and “chart question answering”. Furthermore, the resulting survey articles have been investigated to expand our list of keywords. Through an iterative refinement process, we arrived at the final search string of: “Transformer ("Chart Derendering"$|$"Chart question answering"$|$"Chart Understanding"$|$"Chart captioning"$|$"Chart Reasoning")”. This process involved exploring various combinations of keywords and Boolean operators to ensure comprehensive coverage of relevant literature while minimizing retrieval of irrelevant studies. All searches spanned the period from January 2020 until June 2024 and included journal articles, conference papers, and review papers published in English only arriving at a collection of 227 articles.

The selection criteria were based on the PRISMA Statement~\citep{page2021prisma}. The inclusion and exclusion criteria for this study are outlined as follows.

\paragraph{Inclusion Criteria}
\begin{enumerate}
  \item Focus on frameworks utilizing at least one transformer model in their pipeline.
  \item Focus on the field of computer science.
  \item Publication within the period from 2020 to June 2024.
  \item Preprints available on arXiv\footnote{\url{https://arxiv.org/}}.
  \item Literature in the English language.
\end{enumerate}

\paragraph{Exclusion Criteria}
\begin{enumerate}
  \item Published before 2020.
  \item Solely focusing on data narration through visualization.
  \item Solely focusing on benchmarking commercial Large Language Models (LLMs), Vision-Language Models (VLMs), and Multi-modal Large Language Models (MLLMs).
  \item Solely focusing on the generation of chart images from textual prompts.
  \item Solely focusing on visualization recommendation.
\end{enumerate}

\noindent This study is based only on original research articles, review papers and conference papers. To maintain the quality of the review, all duplications were checked thoroughly. Abstracts of the articles were checked deeply for the analysis of the articles to ensure the quality and relevance of academic literature included in the review process. The initial screening phase resulted in discarding 129 articles keeping 98 articles only. A careful evaluation of each research paper was carried out according to the mentioned criterion. After assessing each article on the aforementioned inclusion and exclusion criteria, The final selection approached 32 articles.

\subsection{Contributions}\label{subsec1-3}
This work presents the following key contributions to the field of CU:
\begin{itemize}
  \item Bridging the knowledge gap for early-career researchers and practitioners in CU by providing a comprehensive survey of key pre-trained transformer models.
  \item Structuring the CU domain taxonomy into three distinct layers: perception, comprehension, and reasoning. This taxonomy is a refinement of existing taxonomies by placing established tasks within these layers and introducing a new task, "chart derendering," resulting in a total of seven CU tasks.
  \item Summarizing and evaluating the current benchmarks and datasets employed in the development of CU frameworks.
  \item Categorizing CU frameworks based on the scope of CU tasks addressed, distinguishing between frameworks that handle single or multiple tasks.
  \item Conducting a comparative analysis between the reviewed frameworks on the benchmarking datasets available.
  \item Discussing the current challenges, potential applications, and future research directions necessary to advance the field of CU.
\end{itemize}

\subsection{Paper Organization}\label{subsec1-4}
This paper is organized into eight sections, providing a comprehensive overview of CU research using transformer models. Section~\ref{sec2} establishes key pre-trained transformer architectures observed to be utilized in multiple CU settings. It covers advancements in transformer models relevant to CU tasks. Section~\ref{sec3} introduces a taxonomy of the fundamental CU tasks including CQA, Chart Derendering, and Chart-to-Text detailing the characteristics of each task. Section~\ref{sec4} describes the datasets utilized across various CU tasks and the available benchmarks. Section~\ref{sec5} reviews frameworks and how the CU tasks developed over time. Section~\ref{sec6} outlines the metrics used to assess the quality of the developed frameworks and provides a comparative analysis of the reviewed frameworks. Section~\ref{sec7} highlights the current challenges that affect CU frameworks and proposes potential directions for improvement, applications and future research. Finally, Section~\ref{sec8} concludes the review.

\section{Background}\label{sec2}

This section provides an overview of the advancements made in transformer models frequently utilized in the CU domain across different modalities including textual, visual and tabular modalities.

\begin{figure}[h]
\centering
\includegraphics[width=0.4\textwidth]{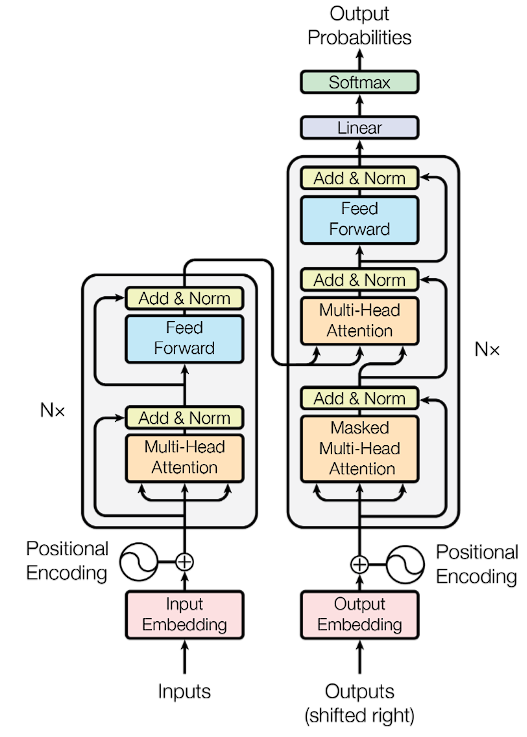}
\caption{The Transformer Model Architecture~\citep{vaswani2017attention}.}\label{fig-ch2-transformer-ov}
\end{figure}

\textbf{Textual Modality.} The original transformer model architecture (i.e. \textit{transformer-ov}), introduced by~\citep{vaswani2017attention}, has seen numerous extensions across various domains, among which is CU. The architecture leverages the attention mechanism which enables models to learn dependencies and contextual relationships within input representations, being text sequences, image patches, or multimodal representation. As illustrated in Figure~\ref{fig-ch2-transformer-ov}, the transformer comprises encoder and decoder blocks commonly employed in natural language processing tasks. The encoder processes input sequences, generating embeddings, while the decoder generates output sequences conditioned on both the encoder's embeddings and its own input. The self-attention mechanism, a key innovation, empowers the model to focus on relevant input components, facilitating the learning of long-range dependencies. In essence, the encoder maps input data to a latent space, while the decoder generates new content based on this latent representation. Encoder and decoder blocks can incorporate multiple attention heads for specialized tasks such as classification or entity recognition.

BART~\citep{lewis-etal-2020-bart} and T5~\citep{raffel2020exploring} are two influential sequence-to-sequence extensions of the \textit{transformer-ov} that are utilized in summarization, QA, and comprehension tasks. They vary across six dimensions: (1) their pretraining objective, (2) the used activation functions, (3) their parameter initialization, (4) the pretraining corpus, (5) how the positional embedding is encoded, and (6) their tokenizers. Changes in these dimensions can be applied to other architectures as well. 

\begin{figure}[h]
\centering
\includegraphics[width=0.7\textwidth]{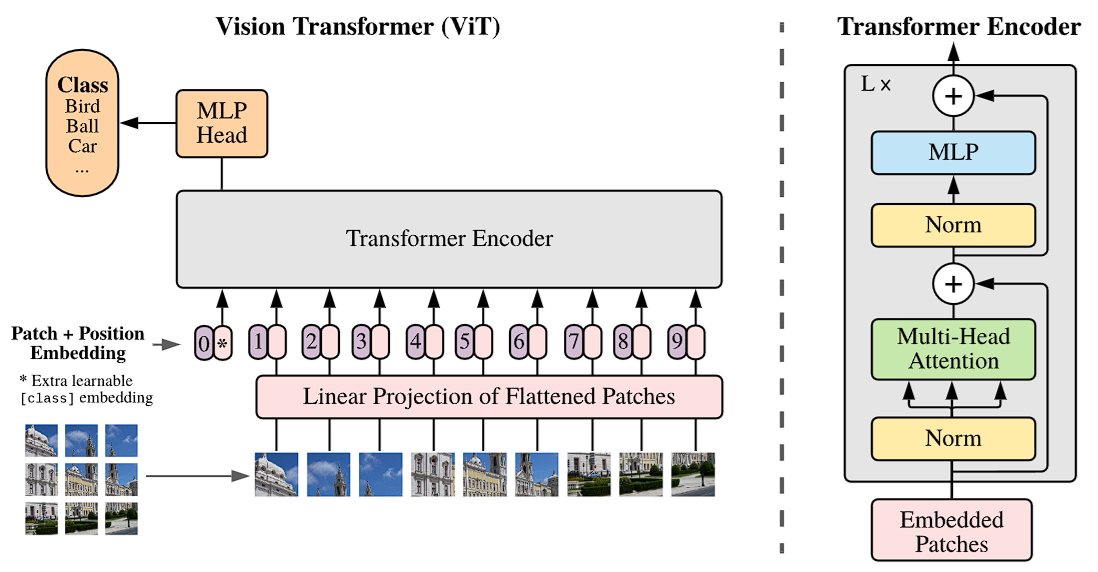}
\caption{The ViT Model Architecture~\citep{dosovitskiy2020image}.}\label{fig-ch2-vit}
\end{figure}

\textbf{Visual Modality.} The adoption of the \textit{transformer-ov} for vision tasks broadened the scope of its applicability with the introduction of the concept of patching images and the Vision Transformer (ViT)~\citep{dosovitskiy2020image}. Once an image is divided into a grid of patches, each patch is then represented as a sequence of pixels, and the transformer is used to learn the relationship between these patches. Hence, by applying ViT or one of its successors to a chart image, we can obtain effective representation and embeddings of the chart components. As shown in Figure~\ref{fig-ch2-vit}, the ViT model utilizes an encoder block with its input sequence being a linear projection of the flattened patches. In contrast to the ViT~\citep{dosovitskiy2020image} architecture, which generates single-resolution feature maps and exhibits quadratic computational complexity with respect to image size, the Swin Transformer~\citep{liu2021swin} introduces a hierarchical feature map structure and linear computational complexity. By partitioning images into local windows and computing self-attention within these windows, the Swin Transformer~\citep{liu2021swin} offers enhanced efficiency and adaptability for various vision tasks. 

Building upon the foundational visual encoders of ViT~\citep{dosovitskiy2020image} and Swin Transformer~\citep{liu2021swin} architectures, researchers have initiated the development of Vision-Language Pretrained (VLP) models. \cite{lee2023pix2struct} introduce Pix2Struct, a pre-trained image-encoder-text-decoder model grounded in the ViT architecture. By incorporating textual components into the input image, the model adapts to diverse visual-language tasks through fine-tuning. To accommodate variable image resolutions, the encoder's input layer is modified, preserving image quality. Pix2Struct offers two variants, with the base model comprising 282 million parameters and a transformer-based decoder characterized by a 768-dimensional hidden size and a 128-token sequence length. Donut~\citep{kim2022ocr} is another VLP specialized for document understanding. It integrates a Swin Transformer~\citep{liu2021swin} as its encoder and a BART~\citep{lewis-etal-2020-bart} decoder. The model comprises 143 million parameters. The benefit of these VLPs is that they eliminate the need for OCR utilization.

\begin{figure}[h]
\centering
\includegraphics[width=0.7\textwidth]{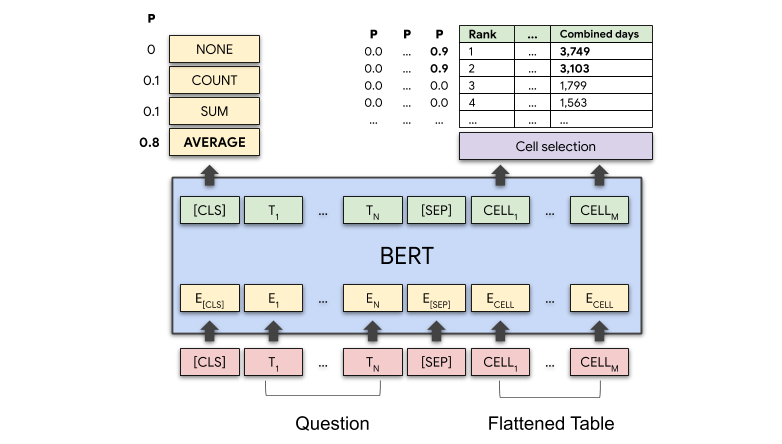}
\caption{The TaPas Model Architecture~\citep{herzig-etal-2020-tapas}.}\label{fig-ch2-tapas}
\end{figure}

\textbf{Tabular Modality.} One of the architectures that plays a significant role in CU tasks is the TaPas architecture, or \textbf{Ta}ble \textbf{Pa}rser~\citep{herzig-etal-2020-tapas}, which was developed to address the Table Question-Answering (Table-QA) task. As certain CU solutions rely on the utilization of tabular data, TaPas facilitates the embedding of the table modality. The architecture (as shown in Figure~\ref{fig-ch2-tapas}\footnote{The architecture was taken from the official blog post in \url{https://blog.research.google/2020/04/using-neural-networks-to-find-answers.html} (accessed on 12/03/2024)}) effectively addresses the Table-QA task by extending the BERT~\citep{devlin-etal-2019-bert} encoder with table structure-based positional embeddings followed by a classification layer for cell selection. BERT itself is an encoder-only transformer block based on the original encoder of~\citep{vaswani2017attention}. Its development focused on defining the Masked Language Modelling (MLM) pre-training objective as it forms bidirectional contexts in its learned attention.

The aforementioned architectures provide the foundation for the recent advancements in addressing the three main modalities in CU tasks, image, text, and the data table, respectively. Combining different modalities through utilizing the transformer architectures opened a door to vast amounts of possibilities. Covering the body of literature across the vision language domain is out of the scope of this research, however, \citep{fields2023vision} provides a survey covering this topic.

\section{Chart Understanding}\label{sec3}
\begin{figure}[h]
\centering
\includegraphics[width=0.8\textwidth]{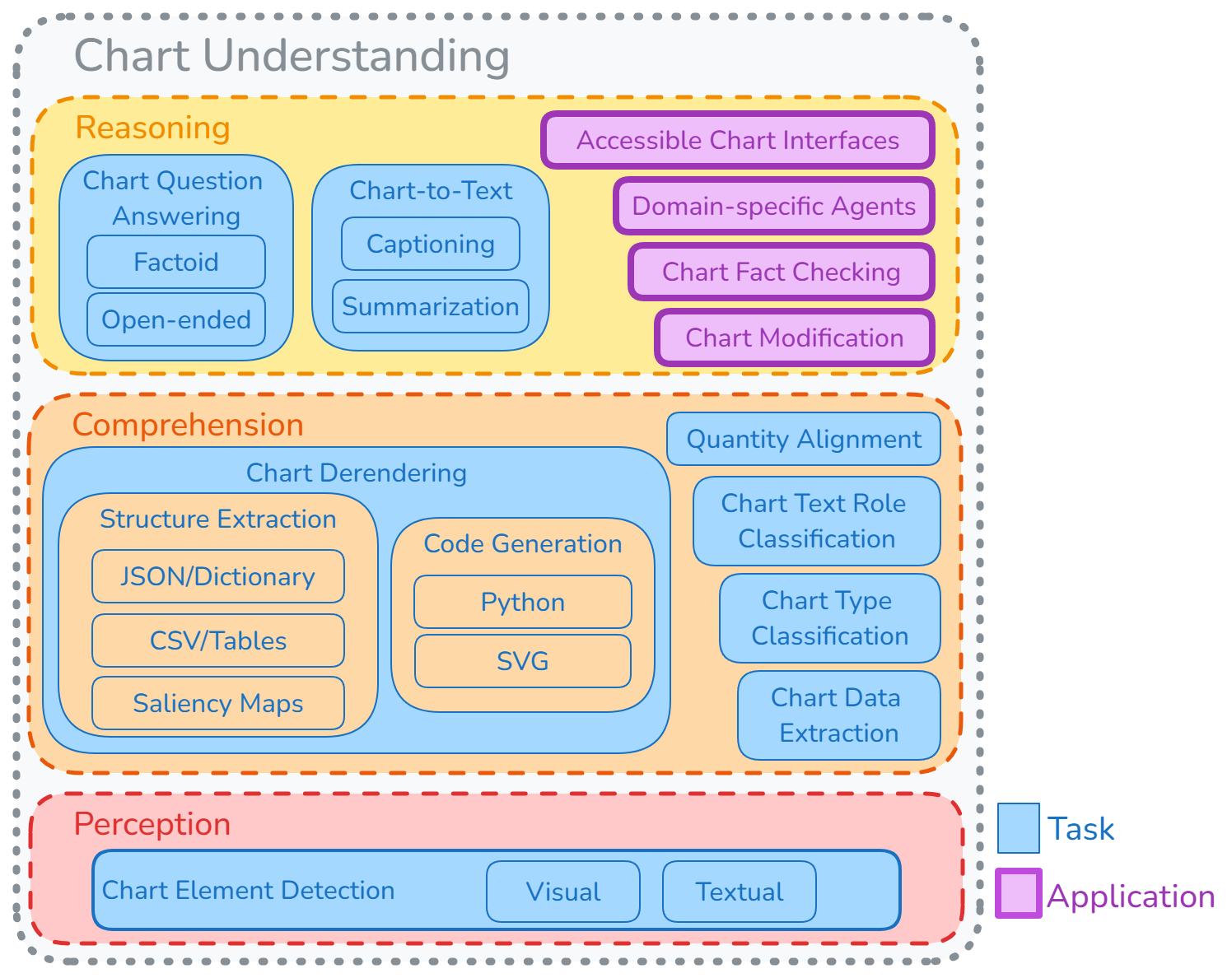}
\caption{Chart Understanding Domain Taxonomy. It is composed of three layers of cognitive tasks: perception, comprehension and reasoning.}\label{fig-ch3-taxonomy}
\end{figure}

Building upon the advancements in transformer architectures across diverse modalities, this section outlines the CU domain, encompassing its core tasks and applications. To establish a common ground, essential terminologies and concepts are introduced. Subsequently, a comprehensive taxonomy of CU tasks is presented, categorized into perception, comprehension, and reasoning layers, as illustrated in Figure~\ref{fig-ch3-taxonomy}. The hierarchical structure mirrors the progressive cognitive demands inherent to each task category, aligned with human cognitive processes when engaging with charts. Table~\ref{tab3-1} gives a summarized review of the frameworks used to solve one or more of the CU tasks while utilizing a transformer architecture for their solution. The present study focuses on CU tasks predicated on the availability of pre-existing chart images, thereby excluding tasks involving chart generation from raw data or textual descriptions.

\begin{table}[h!]
\caption{Overview of CU tasks categorized by cognitive layer, alongside representative frameworks employing transformer architectures.}\label{tab3-1}
\setlength{\tabcolsep}{0.02pt}
\begin{tabular*}{\textwidth}{@{\extracolsep\fill}ccc}
\toprule%
Task & Type & Example Frameworks\\
\midrule
\multicolumn{3}{l}{\textbf{\textit{Layer 1: Perception}}} \\
\multirow{2}{*}{Chart Element Detection}  & Visual  & \citep{lal2023lineformer} \\\cmidrule{2-3}
                      & Textual  & \citep{kim2024text} \\

\midrule
\multicolumn{3}{l}{\textbf{\textit{Layer 2: Comprehension}}} \\
\multirow{2}{*}{Chart Derendering} & Structure Extraction & 
\makecell{
\citep{xia2023structchart, liu-etal-2023-deplot, masry-etal-2023-unichart, han2023chartllama, chen2024onechart, liu-etal-2024-mmc, kim2024simplot}
}
\\\cmidrule{2-3}
                  & Code Generation &
                  \makecell{
                  \citep{liu-etal-2023-matcha, han2023chartllama, masry2024chartinstruct, zhang2024tinychart}
                  }
                  \\\cmidrule{1-3}
                  
Quantity Alignment & -- & \citep{dong2024ttc}\\
Chart Text Role Classification & -- & \citep{kim2024text} \\
Chart Type Classification & -- & \citep{liu-etal-2024-mmc, xia2024chartx} \\
Chart Data Extraction & -- & \citep{liu-etal-2024-mmc, masry2024chartinstruct} \\

\midrule
\multicolumn{3}{l}{\textbf{\textit{Layer 3: Reasoning}}} \\
\multirow{2}{*}{Chart Question Answering}  
& Factoid & \makecell{
\citep{singh2020stl, levy2022crct, masry2021integrating, jain2022tapasqa,
masry-etal-2022-chartqa, wei2024mchartqa, liu-etal-2023-matcha, cheng2023chartreader, 
zhou2023enhanced-chartt5, masry-etal-2023-unichart, han2023chartllama, liu-etal-2024-mmc, 
meng2024chartassisstant, masry2024chartinstruct, xia2024chartx, zhang2024tinychart}
}\\\cmidrule{2-3}
& Open-ended & \makecell{\citep{kantharaj-etal-2022-opencqa, masry-etal-2023-unichart, liu-etal-2024-mmc, meng2024chartassisstant, masry2024chartinstruct, zhang2024tinychart}}\\\cmidrule{1-3}

\multirow{2}{*}{Chart-to-Text}  
& Captioning & \citep{han2023chartllama, xia2024chartx}\\\cmidrule{2-3}
& Summarization & \makecell{\citep{kantharaj-etal-2022-chart, zhou2023intelligent, liu-etal-2023-matcha, 
cheng2023chartreader, zhou2023enhanced-chartt5, masry-etal-2023-unichart, han2023chartllama,
masry2024chartinstruct, zhang2024tinychart}}\\\cmidrule{1-3}

\multirow{4}{*}{Applications}  
& Chart Fact-Checking &
\makecell{
\citep{akhtar2023reading, huang2023lvlms, 
akhtar2024chartcheck, masry2024chartinstruct, krichene2024faithful}
}
\\\cmidrule{2-3}
& Domain-Specific Agents & \citep{kim2024chatgpt} \\\cmidrule{2-3}
& Chart Modification &
\makecell{
\citep{xia2023structchart, han2023chartllama, xia2024chartx, zhang2024tinychart}
}
\\\cmidrule{2-3}
& Accessible Chart Interfaces & 
\makecell{
\citep{choi2022intentable, hsu2024scicapenter, singh2024figura11y, chen2024onechart}
}
\\

\botrule
\end{tabular*}
\end{table}

\subsection{Key Concepts}\label{subsec3-1}
\begin{definition}[Chart Images]
 Images that serve as versatile visual representations of data, finding application across diverse domains. Within academic and research contexts, they facilitate data dissemination and interpretation. In the business and industrial spheres, chart images are instrumental in data-driven decision-making and strategic planning. Additionally, they contribute to knowledge transfer in education, inform public discourse through media, and support governmental policy formulation.
\newline
\end{definition}

\begin{definition}[Optical Character Recognition (OCR)]
 A prevalent component in CU frameworks that enables the extraction of textual content and corresponding spatial coordinates from chart images. To isolate the impact of OCR performance, certain studies employ a perfect "oracle" OCR system, providing ground truth bounding boxes and textual content for chart elements, as a benchmark for evaluation~\citep{singh2020stl, masry-etal-2022-chartqa}.
\newline
\end{definition}

\begin{definition}[Dynamic Encoding]
 Tackles the challenge of vast chart vocabularies by maintaining a local dictionary for each chart, and tracking its text-based components (e.g., x-axis labels). Introduced by~\citep{kafle2018dvqa}, this technique helps structure chart data but struggles with generalizing to misspellings or textual variations (e.g., "United States of America" vs. "USA" are treated differently).
\newline
\end{definition}

\begin{definition}[Pre-training]
 Plays a crucial role in the advancements of transformer models. It involves training the model on one or more tasks to cultivate domain-specific knowledge within its architecture. While requiring significant data and computational resources, pre-training improves model convergence during fine-tuning on various downstream tasks~\citep{kotei2023systematic}.
\end{definition}

\subsection{Layer 1: Perception}\label{subsec3-2}
Chart perception constitutes the initial stage of human-chart interaction, focusing on the immediate visual interpretation of chart elements. This foundational process involves the identification and comprehension of fundamental chart components such as axes, titles, legends, data points, and visual encodings~\citep{wu2021ai4vis}. Accurate detection and recognition of these elements are essential prerequisites for subsequent higher-level chart understanding tasks~\citep{bajic2023review}.

\subsection{Layer 2: Comprehension}\label{subsec3-3}
Chart comprehension constitutes the subsequent phase in chart understanding, building upon the foundational processes of perception. This stage involves the cognitive integration of visually extracted chart elements to derive semantic meaning. Core tasks within chart comprehension encompass the extraction of textual and numerical data, the classification of chart types, the alignment of quantities, and the reconstruction of the underlying data structure through chart derendering. These processes collectively facilitate a deeper understanding of the chart's underlying information and its representation.

\subsubsection{Text Role Classification}\label{subsubsec3-3-1}
Chart text role classification is a task that involves identifying and classifying the semantic roles of textual elements within scientific charts. The goal is to determine the function or purpose of different text components, such as titles, labels, and annotations, within the context of the chart. In the context of scientific charts, text role classification aims to categorize text into specific roles, which may include, as presented in \citep{davila2021icpr, davila2022icpr}: (1) Chart Title: The main title of the chart that describes what the chart represents; (2) Legend Title: The title of the legend that explains the symbols or colours used in the chart; (3) Legend Label: The labels that correspond to the different elements represented in the legend; (4) Axis Title: The titles for the axes that indicate what data is being represented; (5) Tick Label: The labels on the axes that denote specific values or categories; (6) Tick Grouping: Text that groups tick labels together for clarity; (7) Mark Label: Text that differentiates between various marks or data points on the chart; (8) Value Label: Text that displays quantitative data for specific points or areas on the chart; (9) Other: Any other text that does not fit into the above categories. This task enhances the readability and interpretability of scientific charts, as it helps ensure that all necessary elements are present and correctly labelled. Effective text role classification can also support automated systems that provide feedback to authors on the completeness and clarity of their charts \citep{kim2024text}.

\subsubsection{Chart Data Extraction}\label{subsubsec3-3-2}
Chart data extraction constitutes the subsequent phase in chart understanding, focusing exclusively on the semantic interpretation of detected chart markers. This process involves the precise identification and extraction of numerical or categorical data values associated with these markers. Importantly, chart data extraction is bound to the extraction of raw data points, excluding the generation of structured data formats \citep{shahira2021towards, singh2023towards, farahani2023automatic}. For instance, chart data extraction enables the identification of axis ranges, the quantitative values of bar segments or pie slices, among other data elements.

\subsubsection{Chart Type Classification}\label{subsubsec3-3-3}
Chart-type classification is a foundational task in chart understanding, analogous to image classification in computer vision \citep{dhote2023survey}. It involves the categorization of a chart image into predefined classes based on its visual composition. This process necessitates a comprehensive analysis of chart elements, including axes, labels, data points, and their spatial relationships \citep{bajic2023review}.

\subsubsection{Quantity Alignment}\label{subsubsec3-3-5}
The innovative research by~\citep{dong2024ttc} introduces a unique challenge of understanding numerical information presented in various formats—text, tables, and charts. TTC-QuAli~\citep{dong2024ttc} is a groundbreaking resource for linking related quantities across these modalities by introducing a quantity alignment task. Quantity alignment can be considered primarily a comprehension task over charts, as it involves understanding and interpreting the relationships between textual descriptions, tables, and visual elements in charts. Two quantity alignment settings are introduced: single (one-to-one) and composite (one-to-many). In single alignment, each quantity mentioned in the text is linked to a single corresponding cell in a table or a single element in a chart. The focus is on establishing a direct relationship between one specific quantity and one specific target, ensuring a clear and straightforward mapping. For example, if a sentence states, "The sales in Q1 were 100 units," it would link the quantity "100" to a specific cell in a table that represents Q1 sales. Conversely, composite alignment involves linking a single quantity in the text to multiple cells in a table or multiple elements in a chart. It can include scenarios where a quantity may require aggregation (e.g., summing values from multiple cells) or where multiple related quantities are involved in a single statement. For instance, if a sentence states, "The total sales for Q1 and Q2 were 250 units," it would link the quantity "250" to multiple cells representing sales for both Q1 and Q2, potentially requiring the model to sum those values.

\subsubsection{Chart Derendering}\label{subsubsec3-3-4}
\begin{figure}[h]
\centering
\includegraphics[width=1\textwidth]{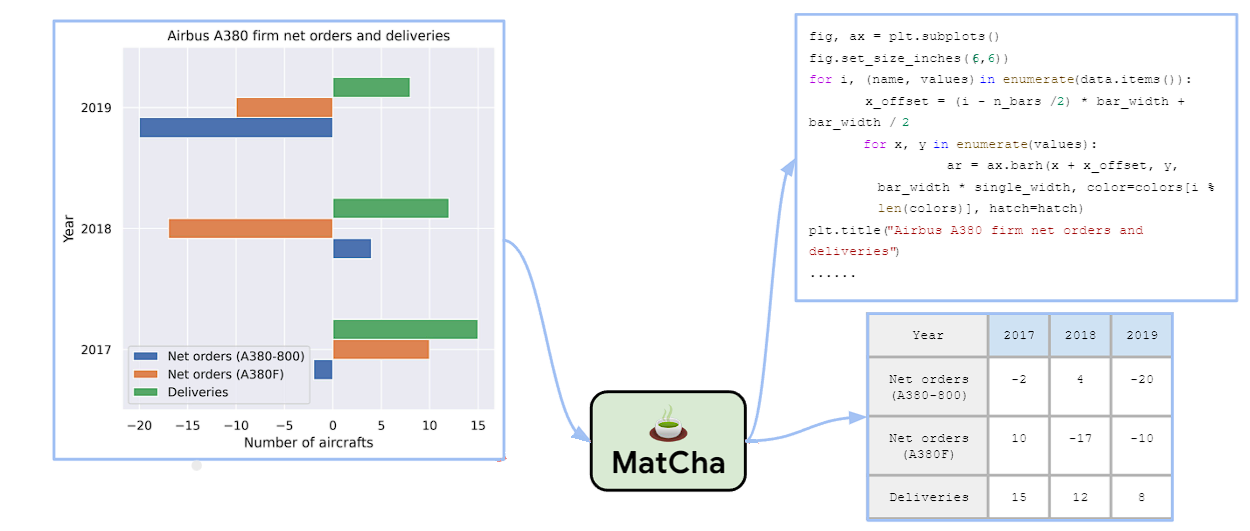}
\caption{An example of the target outputs in the chart derendering task~\citep{liu-etal-2023-matcha}.}\label{fig-ch3-derender-example}
\end{figure}

The term “Chart Derendering” was first introduced by \citep{liu-etal-2023-matcha}, with MatCha being the pioneering framework in this area. MatCha aims to not only recover the underlying data table of a chart but also the code used to generate it. To advance chart understanding and enable models to better comprehend chart layout and visual attributes, chart derendering can be categorized into two primary tasks: transforming the chart image into a structured textual or visual format, or reconstructing the original code. Figure~\ref{fig-ch3-derender-example} displays an example of the outputs of the chart derendering task types.

\textbf{Structure Extraction.} This task focuses on extracting the underlying data from a chart image and representing it in a structured format. This format can include a dictionary, a table, a Vega-Lite JSON specification or a saliency map of the chart's layout. Typically, this process involves combining computer vision techniques with heuristics to extract visual features, which are then incorporated into a transformer model \citep{masry-etal-2022-chartqa, zhou2023enhanced-chartt5}. Recent approaches utilize transformer-based chart encoders to learn from visual embeddings \citep{xia2023structchart, liu-etal-2023-deplot, chen2024onechart, liu-etal-2024-mmc, kim2024simplot}.

\textbf{Code Generation.} The task focuses on recovering the code used to render a chart image. This code can manifest as a Python snippet specific to a library like matplotlib, or an SVG/HTML format. MatCha~\citep{liu-etal-2023-matcha} utilized a training corpus exclusively containing Python-rendered chart images. ChartX~\citep{xia2024chartx} provides the Python code used to generate chart images. To advance this field, research could explore the impact of different code representations and incorporate a deeper understanding of chart image structure into models.

\subsection{Layer 3: Reasoning}\label{subsec3-4}
Chart reasoning is the cognitive process of deriving new insights from visual data presented in a chart. Building upon the foundational tasks of chart perception and data extraction, chart reasoning involves inferring relationships, trends, and patterns within the data. Common chart reasoning tasks include question answering and the generation of textual summaries, requiring models to interpret the chart's underlying meaning and communicate it effectively \citep{huang2024detection}.

\subsubsection{Chart Question Answering}\label{subsubsec3-4-1}
CQA constitutes a specialized domain within the broader Visual Question Answering (VQA) paradigm. While VQA typically involves answering questions based on arbitrary images, CQA focuses on answering questions pertaining to chart visualizations \citep{hoque2022chart}. Distinguishing CQA from general VQA is a requirement for advanced reasoning capabilities, often necessitating mathematical operations to derive accurate responses. To emulate human-like interaction with charts, CQA research has explored methods for machines to comprehend and reason over chart representations. Early works in this field predominantly relied on heuristic approaches~\citep{chaudhry2020leafqa, kim2020answering}, while present-day research leverages transformer architectures, reflecting State-of-The-Art (SoTA) advancements~\citep{singh2020stl, masry-etal-2022-chartqa, wei2024mchartqa}.

Transformer-based frameworks for CQA commonly employ question embedding, textual chart summarization, and potential data augmentation techniques. While these strategies contribute to model performance, reconstructing the underlying data table is pivotal for effective question-answering. Several studies, including~\citep{masry2021integrating, jain2022tapasqa, masry-etal-2022-chartqa, cheng2023chartreader}, have demonstrated the efficacy of this approach. By transforming the reconstructed data table into a sequence and feeding it into a subsequent reasoning module, such as TaPas~\citep{herzig-etal-2020-tapas}, these works have achieved promising results.

\begin{figure}[h]
\centering
\includegraphics[width=1\textwidth]{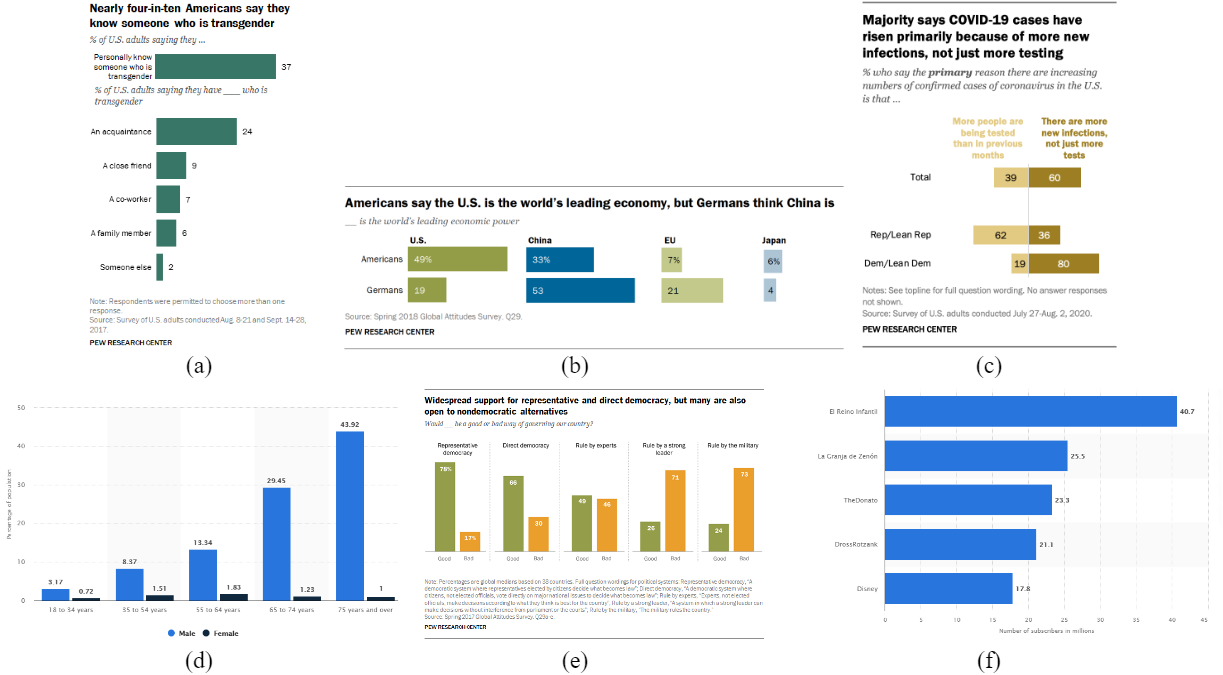}
\caption{Sample chart images from ChartQA and OpenCQA datasets~\citep{masry-etal-2022-chartqa, kantharaj-etal-2022-opencqa}.}\label{fig-ch3-cqa-example}
\end{figure}

CQA tasks are classified into two primary categories based on the nature of the expected answer. Following the nomenclature established in the NLP domain~\citep{kolomiyets2011survey}, questions requiring concise, objective responses are categorized as Factoid CQA, while those demanding a more conversational and comprehensive response are classified as Open-ended CQA. Illustrative examples of each CQA type are presented in Table~\ref{tab3-2}, with corresponding chart images depicted in~\ref{fig-ch3-cqa-example}.

\begin{table}[h!]
\caption{Sample of question types in each CQA task type. Examples are taken from ChartQA and OpenCQA datasets~\citep{masry-etal-2022-chartqa, kantharaj-etal-2022-opencqa}. The “Ref.” column refers to the question’s respective chart images in Figure~\ref{fig-ch3-cqa-example}.}\label{tab3-2}
\setlength{\tabcolsep}{0.5pt}
\begin{tabular*}{\textwidth}{@{\extracolsep\fill}cclc}
\toprule%
\makecell{\textbf{CQA} \\ \textbf{Type}} & \makecell{\textbf{Question} \\ \textbf{Type}} & \textbf{Example} & \textbf{Ref.}\\
\midrule
Factoid & \makecell{Data \\ Retrieval} & How many subscribers did La Granja de Zen3n have? & \makecell{Fig.~\ref{fig-ch3-cqa-example}\\(f)}
\\\cmidrule{2-4}
 & Visual & What is the value of the rightmost light blue bar? & \makecell{Fig.~\ref{fig-ch3-cqa-example}\\(d)}
\\\cmidrule{2-4}
 & Compositional & \makecell[l]{What was the leading YouTube channel in Argentina \\ as of March 2021?} & \makecell{Fig.~\ref{fig-ch3-cqa-example}\\(f)}
\\
\midrule
\makecell{Open\\-ended} 
& Identify & What are the current thoughts on direct democracy? & \makecell{Fig.~\ref{fig-ch3-cqa-example}\\(e)}
\\\cmidrule{2-4}
 & Summarize & \makecell[l]{Explain the distribution of people who know \\ a transgender person?} & \makecell{Fig.~\ref{fig-ch3-cqa-example}\\(a)}
\\\cmidrule{2-4}
 & Compare & \makecell[l]{Compare Americans and Germans views about \\ the world economic leader?} & \makecell{Fig.~\ref{fig-ch3-cqa-example}\\(b)}
\\\cmidrule{2-4}
 & Discover & How do Americans’ see the coronavirus statistics? & \makecell{Fig.~\ref{fig-ch3-cqa-example}\\(c)}
\\

\botrule
\end{tabular*}
\end{table}

\textbf{Factoid CQA.} Datasets consisting of concise answers were considered the pioneering work in the chart understanding domain \citep{kahou2017figureqa, kafle2018dvqa}. Therefore, Factoid CQA was being considered one of the main benchmarking tasks for any multi-modal model in the visual-language domain. The task is concerned with outputting a simple factual answer. The question might require direct data retrieval from the image, information about a visual chart element, or more on the complex nature requiring analytical skills and reasoning to be made to infer the answer. The most recent advancements in solving this task repeatedly utilize a transformer architecture as part of their reasoning component~\citep{singh2020stl, levy2022crct, masry2021integrating, jain2022tapasqa, masry-etal-2022-chartqa, liu-etal-2023-matcha, cheng2023chartreader, zhou2023enhanced-chartt5, masry-etal-2023-unichart, han2023chartllama, wei2024mchartqa, liu-etal-2024-mmc, meng2024chartassisstant, masry2024chartinstruct, xia2024chartx, zhang2024tinychart}.

\textbf{Open-ended CQA.} With the recent emergence of AI systems, such as LLMs, answers with a more open-ended nature desirability are increasing. Furthermore, requests for storytelling are also usually asked in a QA format. Hence, this requires the use of more sophisticated textual decoders than the ones used in the Factoid scenario. The most recent works utilize both extractive and generative models that extend the transformer architecture for their decoders~\citep{kantharaj-etal-2022-opencqa, masry-etal-2023-unichart, liu-etal-2024-mmc, meng2024chartassisstant, masry2024chartinstruct, zhang2024tinychart}.

\subsubsection{Chart-to-Text}\label{subsubsec3-4-2}
Chart-to-text generation surpasses the simple conversion of visual data into textual format. The objective is to produce natural language descriptions or summaries that accurately encapsulate the chart's content. This process enhances accessibility by providing alternative text formats for chart images. Generated text can be categorized as either captions or summaries, each serving distinct communicative purposes. The emphasis lies on creating concise yet informative textual representations. Aligning with \citep{lundgard2021accessible} hierarchical taxonomy of natural language descriptions, captioning primarily focuses on describing visual components and their arrangement. In contrast, summarization encompasses higher-level semantic interpretations, including abstract statistical concepts, perceptual insights, and contextual knowledge.

\textbf{Captioning (i.e., L1~\citep{lundgard2021accessible}).} Often synonymous with generating detailed chart descriptions, this category primarily focuses on extractive text generation. This task demands a model's ability to accurately describe the chart's visual layout and constituent elements, without necessitating complex reasoning. Early approaches to chart captioning relied on template-based methods~\citep{chen2020figure, al2023chartlytics}. In contrast, more recent frameworks utilized transformer-based architectures to generate more sophisticated and informative captions~\citep{han2023chartllama, xia2024chartx}.

\textbf{Summarization (i.e., L2, L3 and L4~\citep{lundgard2021accessible}).} This category implies the generation of abstractive summaries, necessitating advanced reasoning capabilities to produce factually accurate and coherent outputs. Given the abstractive nature of summaries, quantitative metrics are insufficient for a comprehensive evaluation, requiring the incorporation of qualitative measures. Recent advancements in LLMs have led to their integration within more recent frameworks~\citep{kantharaj-etal-2022-chart, cheng2023chartreader, liu-etal-2023-matcha, masry-etal-2023-unichart, zhou2023enhanced-chartt5}. These frameworks prioritize the infusion of chart reasoning capabilities through diverse pre-training strategies.

\section{Datasets}\label{sec4}
The rapid progress in CU is driven by the development of datasets offering multimodal representations of charts and their data. Building high-quality datasets is just as important as improving the design of CU architectures. This section will compare and review relevant datasets used for CU research.

Chart images come in a variety of shapes and visualize different use cases. According to a study done by Datylon\footnote{\url{https://www.datylon.com/blog/types-of-charts-graphs-examples-data-visualization} (accessed on 24/10/2023)}, there are over 60 chart types that exist at the time of submission of this publication. Another important aspect of the chart image is its storage format, which can be in bitmap (.jpeg, .png ), Scalable Vector Graphics (SVG), or defined by a tool (e.g., .json using Vega-Lite). SVG and Vega-Lite formats facilitate accurate reconstruction of the data table as the information is encoded within the image file~\citep{masry-etal-2022-chartqa, masry-etal-2023-unichart}. In contrast, bitmap formats necessitate specific data extraction pipelines.

Some datasets are equipped with a textual corpus that provides question-answer pairs and/or chart summaries. Additionally, some datasets include annotations of the chart elements in the format of bounding boxes. Finally, the ground truth data table could also be included in (.csv) format. Table~\ref{tab4-1} provides an overview of the reviewed datasets in this domain, highlighting their common characteristics and suitability for various CU tasks. The listed datasets in this section are ordered chronologically.

\begin{sidewaystable}
\caption{Summary of the available datasets for addressing chart-related tasks. The abbreviation (GT) stands for Ground Truth. The abbreviation $\text{Img}_f$ stands for Image Format. The symbol \# stands for “Number of”. Numbers are reported in thousands and millions with the notations K and M, respectively. The tasks abbreviations $\text{C}_{qa}$, Tx, Tb, and $\text{O}_{qa}$ stand for Chart Question Answering, Chart to Text, Chart to Table, and Open-ended Question Answering, respectively.}\label{tab4-1}
\setlength{\tabcolsep}{1pt}
\begin{tabular*}{\textheight}{@{\extracolsep\fill}lcccccccccccccccccc}
\toprule%
& \multicolumn{5}{c}{Chart Types\footnotemark[1]} & \multicolumn{3}{c}{Avail. Data\footnotemark[2]} & \multirow{2}{*}{\begin{sideways} $\text{Img}_f$ \end{sideways}} & \multicolumn{2}{c}{Synthetic} & \multirow{2}{*}{\begin{sideways} \#Imgs \end{sideways}} & \multirow{2}{*}{\begin{sideways} \#$\text{QA}_p$ \end{sideways}} & \multirow{2}{*}{\begin{sideways}\#Sum \end{sideways}} & \multicolumn{4}{c}{Task} \\\cmidrule{2-6}\cmidrule{7-9}\cmidrule{11-12}\cmidrule{16-19}%
Dataset & B & S & L & P & $\phi$ & $\text{GT}_{t}$ & $\text{GT}_{s}$ & CA & & Data & Img & & & & $\text{C}_{qa}$ & Tx & Tb & $\text{O}_{qa}$\\
\midrule
\hyperref[subsec4-1]{FigureQA}    & \cmark & \xmark & \cmark & \cmark & \xmark & \cmark & \xmark & \cmark & bitmap & \cmark & \cmark & 140K & 1.55M & -        & \cmark & - & \cmark & - \\
\hyperref[subsec4-2]{DVQA}      & \cmark & \xmark & \xmark & \xmark & \xmark & \cmark & \xmark & \cmark & bitmap & \cmark & \cmark & 300K & 3.4M & -        & \cmark & - & \cmark & - \\
\hyperref[subsec4-3]{ChartBlocks}   & \cmark & \cmark & \cmark & \cmark & \xmark & \xmark\footnotemark[3] & \xmark & \xmark & SVG & \xmark & \xmark & 22.7K & - & -   & - & - & \cmark & - \\
\hyperref[subsec4-3]{Plotly}     & \cmark & \cmark & \cmark & \cmark & \cmark & \xmark\footnotemark[3] & \xmark & \xmark & SVG & \xmark & \xmark & 6.5K & - & -   & - & - & \cmark & - \\
\hyperref[subsec4-4]{PlotQA}     & \cmark & \cmark & \cmark & \xmark & \xmark & \xmark & \xmark & \cmark & bitmap & \xmark & \cmark & 224K & 28.9M & -        & \cmark & - & - & - \\
\hyperref[subsec4-5]{FigCAP (H/D)}  & \cmark & \xmark & \cmark & \cmark & \xmark & \xmark & \cmark & \xmark & bitmap & \cmark & \cmark & 210K & - & 210K        & - & \cmark & - & - \\
\hyperref[subsec4-6]{EC400K}     & \cmark & \xmark & \cmark & \cmark & \xmark & \xmark & \cmark & \cmark & bitmap & \xmark & \xmark & 386K & - & -          & - & - & \cmark & - \\
\hyperref[subsec4-7]{Pew}       & \cmark & \cmark & \cmark & \xmark & \cmark & \xmark & \cmark & \cmark & bitmap & \xmark & \xmark & 9.2K & - & 9.2K        & - & \cmark & - & - \\
\hyperref[subsec4-7]{Statista}    & \cmark & \xmark & \cmark & \cmark & \xmark & \cmark & \cmark & \xmark & bitmap & \xmark & \xmark & 34.8K & - & 34.8K       & - & \cmark & \cmark & - \\
\hyperref[subsec4-8]{ChartQA (H/M)}  & \cmark & \xmark & \cmark & \cmark & \xmark & \cmark & \xmark & \xmark & bitmap & \xmark & \xmark & 4.8K/17K & 9.6K/23K & -    & \cmark & - & \cmark & - \\
\hyperref[subsec4-9]{OpenCQA}     & \cmark & \cmark & \cmark & \cmark & \cmark & \xmark & \cmark & \cmark & bitmap & \xmark & \xmark & 7.7K & 7.7K & 7.7K       & - & \cmark & - & \cmark \\
\hyperref[subsec4-10]{ChartX}     & \cmark & \xmark & \cmark & \cmark & \cmark & \cmark & \cmark & \cmark & bitmap & \cmark & \cmark & 6K & - & -           & \cmark & \cmark & \cmark & \cmark \\
\botrule
\end{tabular*}
\footnotetext[1]{The listed chart types are Bar, Scatter, Line, Pie, and Other.}
\footnotetext[2]{The Available Data can be tables, summary and/or Component Annotations.}
\footnotetext[3]{Data Tables could be reconstructed using the information present in the SVG file.}
\end{sidewaystable}

\subsection{FigureQA}\label{subsec4-1}
The idea behind the creation of FigureQA~\citep{kahou2017figureqa} is to define a CQA corpus that presents CU-specific challenges in addition to VQA challenges.

\textit{Dataset acquisition}. The dataset consists entirely of synthetic data. Both the data tables and the images are synthetically generated. Data tables are sampled from predefined numerical ranges, while images are created through a parameterized process that determines figure type, color scheme, number of elements, and the shapes and quantity of data points. The images are plotted using Bokeh \footnote{\url{https://bokeh.org/}} open-source plotting library. The QA corpus is created from a template-based generation process, with 15 predefined question templates focusing on (yes/no) answers.

\textit{Dataset details}. FigureQA offers train/dev/test subsets across five chart types. It includes a total of 140,000 images with over 1.55 million questions.

\subsection{DVQA}\label{subsec4-2}
To define the challenges of CQA questions, \citep{kafle2018dvqa} introduced DVQA, a dataset exploring CU-specific question types that extend beyond yes/no questions. DVQA is another synthetically generated dataset focusing on bar charts.

\textit{Dataset acquisition}. Focusing on bar charts, three types of underlying data are defined: linear, percentage and exponential. Values for each type are sampled based on heuristics. Matplotlib is then used to generate visualizations with stylistic variations. Following a template-based generation, the QA corpus contains three question types: structure understanding, data retrieval, and reasoning questions. Structure understating questions assess the system’s capabilities of understanding the overall layout, such as the number of bars present, or their orientation (vertical or horizontal). Data retrieval questions focus on the system’s capability of parsing the data values of chart components. Finally, reasoning questions assess the ability to infer new information by performing mathematical operations across multiple chart components.

\textit{Dataset details}. The DVQA dataset consists of 300,000 images and over 3.4 million questions. It provides pre-defined train/dev/test splits for evaluation. Interestingly, there are only 1,576 unique answers across all the questions. This limited vocabulary size within the DVQA dataset necessitates the characterization of its answers being “fixed vocabulary”. Across the three question types, there are 1.6 million reasoning questions, 1.1 million data retrieval questions, and 0.47 million structure understanding questions.

\subsection{Beagle}\label{subsec4-3}
Focusing on crawling chart images in the SVG format, \citep{battle2018beagle} created a dataset of five collections. Each collection relates to the library/database with which the images were created. Although the original corpus was intended to be used in a chart-type classification task, it can still be utilized in chart derendering tasks due to the rich information provided in the SVG format.

\textit{Dataset acquisition}. The authors crawled 20 million web pages with the target of finding SVG-based images. They then extracted a screenshot of the visualization from the web page along with its raw SVG specification. Five visualization tools or databases dominated the crawled SVG specifications. These were then used to identify the different collections offered by the Beagle corpus, namely into D3, Plotly, Chartblocks, Fusion Charts, and Graphiq collections.

\textit{Dataset details}. The Beagle corpus offers over 42,000 SVG-based visualizations. The largest collection is \textbf{Chartblocks} as it comprises 52.4\% of the data, followed by the \textbf{Plotly} collection which comprises 35.7\%. The five collections offer a variety of chart types with D3 offering up to 22 different types. However, the line and bar chart types dominate the five collections, while pie charts are rarely observed and comprise only 5\% of the collected collections.

\subsection{PlotQA}\label{subsec4-4}
To address the limitation of FigureQA and DVQA being entirely synthetic datasets, PlotQA~\citep{methani2020plotqa} leverages real-world data sources. Unlike DVQA, PlotQA introduces more questions that require “open vocabulary” answers. 

\textit{Dataset acquisition}. The acquisition of the dataset followed a four-stage framework. Firstly, online data sources were crawled to curate various data tables. Using real-world data, synthetic plots were generated, although the tool of generation was not disclosed in the original paper. Sample questions were then collected for a subset of 1,400 plots from Amazon Mechanical Turk (AMT)\footnote{\url{https://www.mturk.com/}} workers. Each worker was tasked with writing complex reasoning questions for 5 plots. Finally, after manual analysis of the collected questions, the authors defined 74 question templates. The question types were categorized similarly to those in DVQA.

\textit{Dataset details}. PlotQA offers three chart types and includes 224,377 images with 28.9 million question-answer pairs. Notably, 80.76\% of these questions require “open vocabulary” responses, resulting in a dataset with 5.7 million unique answers.

\subsection{FigCAP}\label{subsec4-5}
For addressing the task of chart captioning (Section~\ref{subsubsec3-4-2}), FigCAP~\citep{chen2020figure} offers two levels of caption details (FigCAP-H and FigCAP-D). FigCAP-H provides high-level chart captions, while FigCAP-D provides detailed captions.

\textit{Dataset acquisition}. Following the generation framework of FigureQA, this dataset contains chart-caption pairs. Ten caption templates are used for generating the captions. The images encompass five chart types: horizontal bar charts, vertical bar charts, line plots, dotted line plots, and pie charts.

\textit{Dataset details}. The total number of chart-caption pairs is 210,024, evenly divided across FigCAP-H and FigCAP-D.

\subsection{ExcelChart400K (EC400K)}\label{subsec4-6}
Since many CU tasks rely on the ability to derender chart images, large datasets that map these images to their corresponding ground truth tables are essential. ExcelChart400K~\citep{luo2021chartocr} offers such a dataset.

\textit{Dataset acquisition}. The process begins by crawling publicly available Excel sheets containing plots. Next, Excel APIs are employed to reconstruct the underlying data tables. However, as the textual chart elements might contain sensitive information, an anonymization process is applied to ensure privacy protection.

\textit{Dataset details}. The dataset is entirely composed of real-world data and encompasses three chart types: bar, line, and pie charts. It contains 386,966 images with their corresponding data tables. Bar charts make up the largest portion of the dataset at 48.4\%, followed by line charts at 31.8\%, and pie charts at 19.8\%.

\subsection{Chart-to-Text}\label{subsec4-7}
Shifting from template-based caption generation, Chart-to-Text~\citep{kantharaj-etal-2022-chart} offers two real-world datasets: Statista and Pew.
\begin{itemize}
  \item \textbf{Statista}
\end{itemize}

\textit{Dataset acquisition}. The dataset is collected from (\url{statista.com}), a website hosting various statistics across diverse topics. A total of 34,811 charts were crawled, along with their data table and human-written descriptions. Leveraging the data table structures, the charts were classified into two categories: simple (with two columns) and complex (with three or more columns). Annotating the charts with their summaries and data tables was a straightforward process, as both elements were readily available. A portion of 20\% of the charts had missing information that the authors manually annotated.

\textit{Dataset details}. Three chart types appear in the dataset: Bar, Line, and Pie. Of the total 34,811 charts, 27,869 are categorized as simple, while the remaining 6,942 as complex. Bar charts make up the largest portion of the dataset at 87.9\%, followed by line charts at 10.2\%.

\begin{itemize}
  \item \textbf{Pew}
\end{itemize}

\textit{Dataset acquisition}. Pew Research (\url{pewresearch.org}) is a website that hosts data-driven articles exploring public opinion on various social issues and demographic trends. The dataset curation included a collection of 3,999 articles, resulting in a corpus of 9,285 chart images. Notably, individual articles may contain multiple chart instances, however, only 143 images had an accompanying data table. In addition to the scarcity of data tables, annotating charts with their summaries was more challenging as paragraphs in the article do not directly refer to the relevant charts. Thus, the annotation was conducted in a three-stage process. First, an OCR was employed in conjunction with a chart-component element classifier to recreate the data tables. Next, a set of heuristics identified a subset of potentially relevant paragraphs that might contain candidate summaries. These heuristics classified the relevance of each candidate paragraph as either "relevant" or "ambiguous." To address the ambiguous cases, AMT workers were tasked with assessing the relevance level of the selected paragraphs.

\textit{Dataset details}. Five chart types appear in the dataset: Bar, Line, Pie, Area, and Scatter plots. Of the total dataset, 1,486 instances represent simple chart structures, while the remaining 7,799 are more complex. Bar charts make up the largest portion of the dataset at 67.9\%, followed by line charts at 26.4\%.

\subsection{ChartQA}\label{subsec4-8}
The ChartQA~\citep{masry-etal-2022-chartqa} dataset, followed the same methodology as PlotQA by utilizing online sources for real-world data tables, yet their generated charts remained synthetic. 

\textit{Dataset acquisition}. To guarantee a broad spectrum of chart topics and styles, four different sources were crawled, i.e., Statista (\url{statista.com}), Pew Research (\url{pewresearch.org}), OWID (\url{ourworldindata.org}), and OECD (\url{oecd.org}). To address the issue of template-based questions, question-answer pairs were collected through two processes. First, human authors created question-answer pairs. Second, machine-generated question-answer pairs were created from human-authored summaries using a pre-trained T5 model (Raffel et al., 2019). This resulted in two subsets: ChartQA-H and ChartQA-M. The focus was on complex reasoning questions. Complex reasoning questions are questions that require different arithmetic and logical operations to reach an answer or may require analysis of different visual attributes of the chart to retrieve the desired values.

\textit{Dataset details}. ChartQA dataset is a large-scale collection of real-world charts and human-authored questions. It includes 9.6K human-authored questions over 4.8K charts, and 23.1K questions generated from human-authored summaries over 17.1K charts.

\subsection{OpenCQA}\label{subsec4-9}
OpenCQA~\citep{kantharaj-etal-2022-opencqa} introduces a new challenge in the CQA task, in which it prompts ChatGPT-like answers.

\textit{Dataset acquisition}. The dataset curates 7.7K humanly authored questions while utilizing real-world charts collected from Pew Research (\url{pewresearch.org}). The question types are categorized into four categories, the \textit{identify} type is essentially a data-retrieval from the chart, \textit{compare} are questions that require analysis over two or more data items or attributes, \textit{summarize} are statistical analysis questions, and \textit{discover} are reasoning questions that require drawing insights.

\textit{Dataset details}. Five chart types appear in the dataset: Bar, Line, Pie, Area, and Scatter plots. Regarding the distribution of question-types, 100 randomly selected questions were analyzed. It was observed that people frequently engage in tasks requiring the \textit{identify} and \textit{compare} of specific chart sections to answer questions.

\subsection{ChartX}\label{subsec4-10}
ChartX~\citep{xia2024chartx} introduces a comprehensive multi-modal evaluation set aimed at rigorously assessing the chart understanding capabilities of MLLMs, addressing the limitations in their performance on complex reasoning tasks involving visual charts.

\textit{Dataset acquisition}. The dataset was synthetically generated through a two-stage pipeline. The first stage focused on creating perception data, including chart images and associated information. The second stage generated cognition data, such as reasoning tasks and redrawing codes. GPT-4~\citep{achiam2023gpt4} was utilized to generate a diverse pool of chart types and topics. For each combination, GPT-4 created chart data, titles, and types. Pre-designed prompt templates guided the generation process, specifying requirements like length and complexity. The second stage centered on creating cognition data for each chart perception sample. Specific instructions and task templates were designed to generate various reasoning tasks, including questions, summaries, and descriptions.

\textit{Dataset details}. ChartX covers 18 distinct chart types, including Bar, Line, Pie, Radar, Bubble, 3D-bar, and Area, providing a diverse set of visual representations. The dataset consists of 6,000 chart images, each accompanied by annotations comprising textual description, CSV-file data file, generating code, and task-specific labels. These labels cover various chart-related tasks, such as identification, comparison, and summarization, totaling 48,000 samples. This comprehensive annotation facilitates a thorough evaluation of MLLMs across multiple chart comprehension and reasoning tasks.

\subsection{Discussion}\label{subsec4-11}
The dataset design significantly influences the development and performance of CU frameworks. The reviewed datasets showcase the vast diversity of chart types and storage formats. This highlights the challenge of developing CU frameworks that can generalize well across a wide range of chart representations. A key aspect is the access to formats that offer direct mappings between the chart image and the underlying data table. SVG and Vega-Lite formats hold a clear advantage in this regard. Since they encode the data within the image, they can potentially facilitate accurate table reconstruction. While each dataset is typically developed with a specific CU task in mind, having element annotations can also be beneficial for chart derendering tasks, regardless of the target downstream task. Datasets not only define benchmarks for each task but also influence pre-training and fine-tuning objectives, fostering the development of models with enhanced reasoning and comprehension capabilities.

\section{Related Work}\label{sec5}
Based on the established taxonomy of CU tasks in Section~\ref{sec3}, this section reviews the frameworks that leverage transformer architectures, building upon those introduced in Section~\ref{sec2}. It also delves into the significance of designing pre-training tasks to equip models with the ability to comprehend charts. Furthermore, it investigates the critical role of training and tuning objectives in equipping models to understand charts and explores the architectural nuances of transformer-based frameworks. The introduced frameworks are categorized based on their task scope: single-task or multi-task within a unified architecture. They are presented chronologically, prioritizing preprints where available.

\subsection{Single-Task Frameworks}\label{subsec5-1}
Given the varying levels of research maturity within the chart understanding domain, this section presents a structured overview. We begin by examining tasks with fewer proposed solutions and then transition to those with a more established research base. 

In order to address the quantity alignment task, \citep{dong2024ttc} extended VL-T5~\citep{cho2021unifying} and VisionTaPas~\citep{masry-etal-2022-chartqa} work to support multiple modalities (text, tables, and charts) through several adaptations in their serialization and encoding processes. The key modifications made to facilitate this multi-modal support are as follows: (1) Sequential Serialization: The input data is serialized in a specific order that reflects their arrangement in documents. Text, tables, and charts are processed sequentially, allowing the model to understand the context and relationships among different modalities; (2) Special Tokens: The model incorporates special tokens to distinguish between different types of inputs. For example, tokens are added to identify text snippets, table cells, and chart elements. This helps the model to recognize and differentiate between the various modalities during processing. These adaptations produce TTC(VL-T5) and TTC(VisionTaPas) that effectively handle the complexities of multi-modal data, allowing them to learn and represent the relationships between text, tables, and charts in a unified manner, which is essential for tasks such as quantity alignment. Additionally, contrastive learning is applied, producing ConTTC(VL-T5) and ConTTC(VisionTaPas), to create a shared semantic space where linked quantities from text, tables, and charts are positioned closer together, while unlinked quantities are pushed apart. This approach enhances the models’ ability to accurately align quantities across different modalities by leveraging contextual information and improving its generalization across diverse datasets.

Drawing inspiration from multimodal document layout solutions, \citep{kim2024text} address the text role classification task by proposing a method that fine-tunes two pretrained multimodal document layout analysis models, LayoutLMv3~\citep{huang2022layoutlmv3} and UDOP~\citep{tang2023unifying}, specifically for classifying text roles in scientific charts. The models utilize three modalities of input: text elements, chart images, and layout information (bounding box coordinates of the text). This multimodal approach allows the models to leverage both visual and textual information, which is crucial for understanding the context of the text within the charts. The impact of data augmentation and balancing techniques on model performance is investigated to improve the robustness and effectiveness of the models, especially when dealing with limited training data. The models are evaluated on various chart datasets, including ICPR22~\citep{davila2022icpr}. The evaluation includes measuring performance metrics such as the F1-macro score, which assesses the models' ability to classify text roles accurately across different classes. The robustness of the models is tested on a synthetic noisy extension of ICPR22 to evaluate how well they perform under challenging conditions. Overall, the combination of multimodal inputs, fine-tuning of advanced models, and the application of data augmentation and balancing techniques contribute to effectively solving the text role classification task in scientific charts. The results indicate that LayoutLMv3 outperforms UDOP and demonstrates robustness and generalizability across various datasets.

\subsubsection{Chart Question Answering}\label{subsubsec5-1-1}
Early CQA approaches~\citep{kafle2018dvqa, chaudhry2020leafqa} leveraged VQA techniques, ignoring the importance of the chart components' structure. To address this issue, STL-CQA~\citep{singh2020stl} proposed the first, to our knowledge, transformer-based framework that leverages the structural properties of charts. The framework comprises three modules, with a dedicated reasoning module employing a multi-layer transformer architecture. This module is trained on specific tasks to extract and comprehend underlying chart properties. The framework leverages chart images and input questions without requiring data table reconstruction. Both localization and encoding modules construct featurized vectors for the chart image and question, respectively. Localization highlights the key chart components by using a Mask R-CNN~\citep{he2017mask} object detection model. Encoding, on the contrary, maps the input question text with the respective chart components’ tokens by using the dynamic encoding model introduced by~\citep{kafle2018dvqa}. Within the reasoning module, two transformer blocks act as encoders. The first captures the relationship between chart elements, while the second focuses on the input question. Finally, a cross-attention transformer block leverages the encodings from the previous blocks and outputs the answer using a classification head. Each transformer block has a set of defined pre-training tasks as follows, (1) the Chart Relation Encoder was pretrained on chart structure tasks, to learn the types of chart elements, their positions, and their attributes such as color and pattern. (2) the Question Encoder was trained with the MLM task for the questions, focusing on the chart vocabulary or words that alter the meaning instead of random masking. (3) the Cross-modality transformer block was pre-trained on a task similar to BERT’s~\citep{devlin-etal-2019-bert} next-sentence prediction. Yet, the original sentence is replaced with a mismatched sentence while the classifier learns to identify it. One shortcoming of this framework is its treatment of the question-answering task as a fixed-vocabulary classification problem. Additionally, it relies on the assumption of having a text oracle.

Another framework focusing solely on the CQA task is the CRCT~\citep{levy2022crct} framework. In an attempt to expand the answers to be of an open-vocabulary nature, a co-attention transformer with a classification-regression head was developed. The CRCT framework has two main modules: a detection module, followed by a question-answering module. The detection module takes that chart and question, and extracts visual and textual encodings, respectively. The chart components are detected using the object instance segmentation model, Mask R-CNN~\citep{he2017mask}. Textual encodings are a product of a pre-trained BERT~\citep{devlin-etal-2019-bert} that combines the raw question text and the extracted textual components from the chart image extracted using an OCR. The question-answering module implements a co-attention architecture that works on the extracted visual and textual encodings, followed by a hybrid-prediction head that classifies the type of question and infers a textual or numerical answer. Unlike STL-CQA, the co-attention transformer block was not pre-trained on a separate task, and the whole framework was trained End-to-End (E2E) on the CQA task only. While performing well on PlotQA structural questions, the model exhibited limitations in handling questions requiring mathematical reasoning and analytical skills. Regarding data retrieval questions, the results suggested the need for further training, particularly on chart attributes like element colors.

While the previous two frameworks attempted to solve the CQA task without explicitly retrieving the underlying data table, another body of research solely focuses on chart data extraction, which could then be adapted to solve various CU tasks. To address this, \citep{masry2021integrating} attempted to reconstruct the data table before solving the question-answering task. The first version of their framework, developed as part of CQAW\footnotemark{}, utilized a set of heuristics and computer vision techniques for data recovery. This was followed by the application of TaPas~\citep{herzig-etal-2020-tapas}, as the table question-answering model, which is similar to the work of~\citep{jain2022tapasqa}. The framework was evaluated on FigureQA and Chart Question Answering Challenge (CQAC)\footnotemark[\value{footnote}].

\footnotetext{\url{https://cqaw.github.io/challenge}}

Following the success of applying the TaPas model to the CQA task, \cite{masry-etal-2022-chartqa} resumed their exploration of transformer architectures. They define ChartQA as a new challenging dataset (see Section~\ref{subsec4-8}), and also conducted a comparative study on four transformer-based solutions, two of which combine embeddings of both the visual features of the chart image and its underlying data table. Their work showed promising results, but there was still room for improvement over the newly developed dataset. It presents new challenges with complex reasoning questions that require the aggregation of different mathematical functions to reach an answer.

The ChartQA~\citep{masry-etal-2022-chartqa} framework offers two settings. the first setting assumes the Ground Truth data table is available, while the other reconstructs the data table using their extended version of ChartOCR~\citep{luo2021chartocr}. With the chart components' visual features from Mask R-CNN, the question, and the data table, four models were tested for the CQA module. Two of these models, T5~\citep{raffel2020exploring} and TaPas~\citep{herzig-etal-2020-tapas}, worked with the textual and tabular modalities respectively by processing the question and a flattened version of the table. The other two models, VL-T5 and VisionTaPas, aim to utilize the visual modality in addition to the previous two, allowing them to handle reasoning questions that require visual understanding. VL-T5~\citep{cho2021unifying}, is a vision-language extension of T5, while VisionTaPas, is an extension of TaPas that handles visual features . Being their novel work, VisionTaPas utilizes a ViT~\citep{dosovitskiy2020image} encoder for the chart images and a TaPas encoder for the flattened table and question. Then the encodings are passed to a novel cross-modal encoder. All the utilized transformers were initialized with their pre-trained weights except for the cross-modal encoder, which was trained E2E on the CQA task. One of the limitations of this framework is its dependence on the OCR quality during the data extraction module. Additionally, the new challenging dataset of ChartQA introduced more complex reasoning questions opening the door for research focused on inducing mathematical and analytical skills in future frameworks.

So far, these previous works in CQA research have focused on datasets that restrict answers to factual information (factoid-only). Following the trend established earlier, \cite{kantharaj-etal-2022-opencqa} explore various SoTA architectures to provide a baseline to their newly developed dataset; OpenCQA for open-ended questions. To address this task, they explored three input settings: (1) the chart image and an accompanying article are provided, (2) the chart image and only relevant paragraphs are provided, and (3) only the chart image is provided. In settings (1) and (2), frameworks of extractive or generative nature were employed, while in (3) only generative models were employed. For the extractive framework, BERTQA~\citep{chadha2019bertqa} and ELECTRA~\citep{clark2020electra} were utilized. While the generative models were GPT-2~\citep{radford2019gpt2}, BART~\citep{lewis-etal-2020-bart}, T5~\citep{raffel2020exploring}, VL-T5~\citep{cho2021unifying} and CODR~\citep{prabhumoye2021codr}. The reported limitation of the aforementioned frameworks is their inability to produce answers that address complex logical and arithmetic reasoning.

Complex reasoning remains a challenging task in chart question-answering. To address this, mChartQA~\citep{wei2024mchartqa} proposes a two-stage training strategy aimed at enhancing model comprehension through visual-language alignment and reasoning tasks. mChartQA is a multimodal framework comprising four components: (1) a vision encoder using the CLIP model~\citep{radford2021clip}, (2) a connector aligning visual features with the subsequent LLM text encoder, (3) a Chart-to-Table Engine using DePlot~\citep{liu-etal-2023-deplot}; (4) an LLM, either Qwen~\citep{bai2023qwen} or InternLM2~\citep{cai2024internlm2}. The resulting models, $\text{mChartQA}_{\text{Qwen}}$ and $\text{mChartQA}_{\text{Intern-LM2}}$, demonstrate performance differences, with the latter surpassing the former on evaluated benchmarks.

\subsubsection{Chart-to-Text}\label{subsubsec5-1-2}
Parallel to the advancements in the CQA domain, Chart2text~\citep{obeid2020chart2text} explores the construction of a chart-to-text framework that steps away from template-based Natural Language Generation (NLG). It utilizes a Data-to-Text extension of the architecture proposed by~\citep{gong2019enhanced}. Due to the limited training corpus in terms of the size of samples, Chart-to-Text~\citep{kantharaj-etal-2022-chart} curated a much bigger corpus, (see Section~\ref{subsec4-7}). They also report the performance of various baseline models across image captioning and Data-to-Text tasks. Focusing on the Data-to-Text baseline models as they utilize transformers, the authors evaluated four pre-trained models: Chart2text, Field-Infusing Model, BART and T5. The models were evaluated in three settings that differ based on the input sequence, where the models were given: (1) ground truth table only, (2) reconstructed data tables using ChartOCR extracted data~\citep{luo2021chartocr}, and (3) OCR-only data (bounding boxes and/or the extracted text). While generating textual output, these models exhibited limitations in handling complex patterns and hallucinations. Whereas~\cite{zhou2023intelligent} proposes an encoder-decoder framework for generating chart descriptions focusing on significant data features or patterns, including extreme values, trends, outliers and clusters. Their encoder utilizes a saliency detection model to generate explainable feature maps, a novel approach for transforming chart images into semantically meaningful representations. This framework comprises a visual insight detection module and a chart text generation module, with the latter employing a transformer decoder~\citep{vaswani2017attention}.

\subsubsection{Chart Derendering}\label{subsubsec5-1-3}
The current single-task frameworks predominantly focus on converting chart images into text-based structured representations such as tables or dictionaries~\citep{liu-etal-2023-deplot, xia2023structchart, chen2024onechart, kim2024simplot}. \cite{liu-etal-2023-deplot} explore an OCR-free framework with DePlot. It builds upon MatCha~\citep{liu-etal-2023-matcha}, a pre-trained model, by fine-tuning it specifically for the Chart-to-Table task. This approach achieves high accuracy in table reconstruction. However, DePlot focuses solely on this task and does not address other CU tasks, such as CQA. In order to utilize DePlot in other CU tasks, it employs a separate plug-and-play LLM that solely combines in its input prompt the extracted table and the user query, neglecting the visual information from the original chart image. In essence, DePlot excels at chart-to-table conversion but disregards the valuable visual features present in the chart itself.

Another solution addressing the task is StructChart~\citep{xia2023structchart} which employs an image encoder and text decoder to convert chart images into Linearized Comma-Separated Values Tokens (LCT). Both the encoder and decoder leverage ViT~\citep{dosovitskiy2020image}, with modifications to handle variable resolutions. LCT's sensitivity to entity positional variations motivates the introduction of Structured Triplet Representations (STR), capturing entity relations within the chart. An STR entry comprises three elements: the n-th row header, the m-th column header, and the cell value. Like~\citep{liu-etal-2023-deplot}, the transformed representation serves as input prompts for subsequent LLMs. To augment training data, StructChart generates a synthetic corpus, SimChart9K, using GPT-3.5~\citep{brown2020gpt}, comprising images and corresponding CSV information.

OneChart~\citep{chen2024onechart} introduces a novel approach to chart comprehension by transforming chart images into a structured Python dictionary format, thereby enhancing efficiency and accuracy in extracting underlying chart information. To achieve this, OneChart adopts a newly introduced VLM, Vary-tiny~\citep{wei2024vary-tiny}, which utilizes a SAM-base~\citep{kirillov2023SAM} encoder and an OPT-125M~\citep{zhang2022opt} decoder. Additionally, OneChart incorporates an optional auxiliary decoder consisting of three MLP layers to improve the accuracy of numeric data extraction. OneChart significantly advances chart structure extraction by achieving superior performance while maintaining computational efficiency through a minimized model parameter count.

To address the limitations of the Deplot model~\citep{liu-etal-2023-deplot}, which struggles to effectively leverage the textual context of the chart elements and incorporates only tabular data, \cite{kim2024simplot} introduce SIMPLOT, a model-agnostic approach for accurate chart-to-table conversion. SIMPLOT employs novel methodologies to enhance model comprehension of textual context within charts and isolates essential chart components through a contrastive learning strategy. Additionally, a novel prompting technique coupled with a Large Multimodal Model (LMM) is utilized to foster a more holistic understanding of chart information. SIMPLOT is an encoder-decoder architecture trained in two stages using ViT~\citep{dosovitskiy2020image}. In the first stage, the model, composed of a teacher encoder ($Enc_{chart}^{teacher}$) and a table decoder ($Dec_{table}$) initialized with DePlot, is fine-tuned to convert simple chart images into corresponding tables. Subsequently, knowledge distillation and contrastive learning are applied with an additional student encoder ($Enc_{chart}^{student}$). The goal is to enable the student encoder ($Enc_{chart}^{student}$) to project any original chart into the latent space of simplified chart images. This would enhance the accuracy of generated tables while preserving a minimal total number of model parameters.

\subsection{Multi-Task Frameworks}\label{subsec5-2}
Building upon the foundational research in single-task chart understanding, the subsequent section delves into frameworks designed to address multiple downstream tasks within the domain. These solutions predominantly leverage pre-trained LMMs or VLP models as their backbone. Research efforts primarily converge on two primary strategies: (1) extending pre-training to encompass a broader spectrum of chart understanding tasks or (2) refining pre-trained models through techniques such as prompt engineering and the development of specialized training datasets with an emphasis on instruction tuning or program-of-thought paradigms.

\subsubsection{Pre-training-based}\label{subsubsec5-2-1}
A subset of multi-task frameworks experiments with various CU-focused pre-training tasks. These frameworks aim to enhance model generalization and robustness by exposing the model to a diverse range of chart-related data during the pre-training phase. By pre-training on a combination of low-level perception and comprehension chart understanding tasks, these models seek to acquire a deeper and more comprehensive representation of chart visual and textual information, thereby improving their performance on downstream tasks.

Previous work primarily focused on building frameworks for individual CU tasks. However, many researchers started utilizing the VLP model, Pix2Struct~\citep{lee2023pix2struct} introducing subsequent checkpoints to progressively improve its performance and tackle various CU tasks within a single, E2E framework. To achieve this goal, MatCha~\citep{liu-etal-2023-matcha} adapted the pre-trained Pix2Struct-Base by introducing two additional pre-training tasks. This allows it to address four downstream CU tasks: Chart-to-Table, Chart-to-Code, CQA, and Chart-to-Text Summarization. Specifically, \textit{Chart Derendering} and \textit{Math Reasoning} pre-training tasks were introduced to infuse the model with layout understanding, data table reconstruction, and basic mathematical skills. The \textit{Chart Derendering} task extracts a textual representation of the tabular data and the Python code segment that could have been used to generate an input chart image. \textit{Math Reasoning} was introduced by pre-training the model on textual math reasoning datasets, equipping it with basic mathematical skills. Despite that, the model's ability to address reasoning questions involving visual elements remains limited. Regardless, it has provided a stepping stone to start developing OCR-free frameworks, mitigating the data extraction errors that are common in OCR-dependent frameworks.

Building upon MatCha's success, ChartReader~\citep{cheng2023chartreader} introduces a novel CU vision-language transformer model. This model tackles three tasks: Chart-to-Table, CQA, and Chart-to-Text generation. Their framework consists of two key modules, a transformer-based Chart Component Detection (CCD) module, followed by extending a pre-trained vision-language model for Chart Derendering \& Comprehension (CD\&C). The CCD module eliminates the need for an OCR and performs three stages on the input chart image: (1) Center/Key-point Detection using an Hourglass Network~\citep{newell2016hourglass}, (2) Center/Key-point Grouping using the attention mechanism to obtain embeddings for each centre and key-point, and (3) Component position/type prediction by fusing the learned embeddings and passing them by an MLP layer. The CCD module is pre-trained on a Chart-to-Table task using the EC400K dataset. Taking the resulting embeddings of the chart data as input, the CD\&C module treats Chart-to-(Table/Text) tasks as CQA tasks. It implements a Bidirectional Encoder and Decoder (T5 or TaPas) architecture. The entire E2E framework is then fine-tuned on the CQA and Chart-to-Text tasks using (FQA, DVQA, and PlotQA) and Chart-to-Text, respectively. While ChartReader achieves promising results on all target tasks, there is still room for further optimization. Notably, the reported Chart-to-Table results lack direct comparison with DePlot's evaluation methodology.

ChartT5~\citep{zhou2023enhanced-chartt5} investigates the effectiveness of pre-training tasks specific to CU tasks. These tasks aim to teach the model to map chart images into tables. The researchers introduced two table-masking pre-training tasks: Masked Header Prediction (MHP) and Masked Value Prediction (MVP). These tasks follow the MLM paradigm established by BERT~\citep{devlin-etal-2019-bert}. The intuition behind having two separate masking tasks, unlike traditional MLM, is that recovering a masked table header requires the model to focus on the text present in the chart layout while recovering masked values necessitates mathematical reasoning based on the chart components. Therefore, these tasks train the model for both layout understanding and the ability to perform arithmetic operations. The implemented transformer architecture extends VL-T5. Each modality (chart image, text, and masked data table) is encoded using a specific technique. These encoded representations are then concatenated and fed into a multi-layer encoder. The visual features encoding of the chart image is obtained by using the pre-trained Mask RCNN utilized in~\citep{masry-etal-2022-chartqa}. An embedding of the textual components obtained via an OCR is produced using an OCR encoder. Lastly, the third modality, being a flattened version of the masked data table, is encoded through a shared word embedding layer. A reported limitation of ChartT5 is that the OCR encoder, relying on potentially noisy OCR output, can lead to inaccurate predictions. Additionally, the model struggles with complex reasoning questions. Future work could involve pre-training on tasks that require mathematical reasoning, similar to MatCha.

Finally, UniChart~\citep{masry-etal-2023-unichart} is another OCR-free framework that addresses four CU downstream tasks: CQA, Chart-to-Text Summarization, Chart-to-Table conversion, and OpenCQA. It achieves this by introducing chart-related pre-training tasks that "inject" CU knowledge into the model. These pre-training tasks are categorized into low-level and high-level tasks. The low-level tasks focus on high-level data table reconstruction. They include an explicit task for generating a flattened data table and a data value estimation task. To prepare the model for the downstream CQA, a \textit{Numerical} \& \textit{Visual Reasoning} task was introduced. This task teaches the model to perform basic math operations on 90 templates of QA pairs of the most common mathematical and logical operations. To address OpenCQA, a synthetic corpus of open-ended questions was generated from chart summaries, and the sentence from the summary serves as the target answer. The framework adapts an image-encoder-text-decoder transformer architecture. The image-encoder utilizes Donut~\citep{kim2022ocr}, eliminating the need for an OCR module. BART~\citep{lewis-etal-2020-bart} is used as the text-decoder due to its strength in textual content generation. While UniChart outperforms MatCha on CQA, Chart-to-Text Summarization, and Chart-to-Table tasks, it suffers from hallucinations with densely populated charts and could still be improved for complex numerical reasoning QA.

\subsubsection{Prompt-engineering-based}\label{subsubsec5-2-2}
Complex CU tasks that necessitate intricate reasoning processes remain a formidable challenge for language models of varying scales. To address this, CU-based prompt engineering techniques have emerged as promising avenues of exploration. To address this limitation, researchers have increasingly turned to prompt engineering techniques to enhance model capabilities. Among the prominent approaches are chain-of-thought~\citep{xu2023chartbench, meng2024chartassisstant, masry2024chartinstruct}, program-of-thought prompting~\citep{zhang2024tinychart}, and instruction tuning~\citep{han2023chartllama, liu-etal-2024-mmc, masry2024chartinstruct, meng2024chartassisstant, xia2024chartx}. These methods aim to guide models towards more deliberate and structured reasoning by providing explicit prompts or examples that demonstrate the desired thought process. Chain-of-thought focuses on cultivating a model's ability to break down complex problems into intermediate steps, thereby fostering logical reasoning~\citep{wei2022chain-of-thought}. Program-of-thought extends this concept by generating code-like expressions to solve problems computationally~\citep{chen2022program-of-thought}. In contrast, instruction tuning directly aligns the model's behaviour to specific tasks through exposure to paired instructions and desired outputs, aligning the model’s behaviour with the user’s expectations~\citep{zhang2023instruction-tuning}.

Instruction tuning and supervised fine-tuning diverge in their training data composition. Supervised fine-tuning employs input-output pairs to adjust model parameters, aligning predictions with target outputs. Conversely, instruction tuning introduces an additional layer of complexity by incorporating explicit instructions within the training data, enabling the model to learn to follow diverse task specifications and generalize more effectively across various prompts~\citep{zheng2023learn-beyond-finetune}. The primary factor differentiating various fine-tuning approaches is the dataset's sample properties. Consequently, dataset acquisition becomes crucial. \cite{zhang2023instruction-tuning} categorize dataset acquisition into human-crafted or synthetically generated methods. Synthetic datasets often leverage foundation models like GPT-3.5~\citep{ouyang2022training-gpt3-5} and GPT-4~\citep{achiam2023gpt4}.

The CU frameworks that employ instruction-tuning typically define a synthetic data generation pipeline. In this pipeline, most of the chart data is sourced from a foundation model. For example, ChartAst~\citep{meng2024chartassisstant} utilizes GPT-3.5~\citep{ouyang2022training-gpt3-5} to construct their instruction-tuning set; ChartSFT. The generation of ChartSFT is composed of three stages: (1) table generation, (2) chart generation, and (3) instruction data generation. Similarly, ChartLlama~\citep{han2023chartllama} also has the same three stages. Contrarily, both MMC~\citep{liu-etal-2024-mmc} and ChartLlama~\citep{han2023chartllama} utilize GPT-4~\citep{achiam2023gpt4}. These frameworks also introduce under-explored chart types, such as heat maps and candlesticks. MMC~\citep{liu-etal-2024-mmc} instructions cover a range of comprehension and reasoning tasks, encompassing nine CU tasks. These tasks include chart type classification, chart question answering, and chart data extraction. ChartLlama covers six CU tasks and an application, namely, text-to-chart generation. Subsequently, ChartInstruct~\citep{masry2024chartinstruct} utilizes the APIs of both GPT-3.5 and GPT-4. The choice between GPT-3.5 and GPT-4 depends on the complexity of the task. Tasks requiring complex reasoning, such as chain-of-thought reasoning, are best suited for GPT-4. On the other hand, GPT-3.5 is more effective for tasks with moderate complexity.

To leverage previously developed instruction-tuning datasets, several frameworks employ LMMs. \citep{han2023chartllama, liu-etal-2024-mmc, masry2024chartinstruct, meng2024chartassisstant} illustrate this approach. MMCA~\citep{liu-etal-2024-mmc} utilizes mPLUG-Owl~\citep{ye2023mplug-owl}, which incorporates CLIP~\citep{radford2021clip} for vision encoding and Vicuna~\citep{vicuna2023} as the language model. Similarly, ChartLlama employs LLaVA-1.5~\citep{liu2024improved-llava-1-5} also utilizing CLIP and Vicuna. While both LLaVA-1.5 and mPLUG-Owl share CLIP as the vision encoder and Vicuna as the language model, their architectural approaches diverge. LLaVA-1.5~\citep{liu2024improved-llava-1-5} adopts a monolithic architecture, directly fusing visual and textual representations through an MLP connector. Conversely, mPLUG-Owl~\citep{ye2023mplug-owl} employs a modular design, introducing a visual abstractor module to process visual information prior to integration with the language model. These architectural distinctions contribute to the unique strengths and limitations of each model, influencing their overall performance on a variety of multi-modal tasks.
 
 ChartAst~\citep{meng2024chartassisstant} presents two model variants: ChartAst-D and ChartAst-S. ChartAst-D builds upon the Donut architecture~\citep{kim2022ocr} with 260 million parameters, while ChartAst-S is based on the SPHINX architecture~\citep{lin2023sphinx} with a significantly larger parameter count of 13 billion. Despite its smaller parameter size, ChartAst-D exhibits competitive performance on benchmark datasets, demonstrating comparable results to the larger ChartAst-S model. 

ChartInstruct~\citep{masry2024chartinstruct} offers two system settings and utilizes LLaVA~\citep{liu-2023-llava}, the predecessor of LLaVA-1.5. Unlike MMCA and ChartLlama, which employ CLIP for vision encoding, ChartInstruct adopts UniChart, specifically designed for chart images. The model offers two variants: one integrating Flan-T5~\citep{chung2024-flan-t5} with 3 billion parameters, and another incorporating Llama2~\citep{touvron2023llama} with 7 billion parameters. Ablation studies highlight the importance of chain-of-thought reasoning and chart derendering into code for enhancing the framework's reasoning capabilities.

ChartVLM~\citep{xia2024chartx} employs a cascaded decoder architecture, comprising a base decoder for structural data extraction and a secondary decoder for complex reasoning tasks. This hierarchical approach enables the model to initially extract relevant numerical and textual information from charts before proceeding to higher-order reasoning processes, such as question answering and summarization. An instruction adapter dynamically selects the appropriate task decoder based on user prompts. Trained on a dataset of user instructions and task labels, this adapter achieves 100\% accuracy on the validation set, demonstrating its effectiveness in handling diverse user queries. Comparative evaluations highlight ChartVLM's superior performance in both basic data extraction and advanced reasoning tasks relative to existing models. The model exhibits enhanced capabilities in interpreting chart-specific features and faster inference times.

\cite{zhang2024tinychart} presents TinyChart, an efficient architecture for chart understanding, utilizing a modified ViT~\citep{dosovitskiy2020image} with a Visual Token Merging module. This module aggregates similar visual tokens to reduce the length of visual feature sequences, enabling the model to efficiently process high-resolution chart images. TinyChart employs a Program-of-Thoughts (PoT) learning strategy, which trains the model to generate Python programs for numerical computations, thereby enhancing its ability to answer questions that require mathematical reasoning. With only 3 billion parameters, TinyChart achieves SoTA performance on various chart understanding benchmarks, outperforming larger models like ChartLlama and demonstrating superior inference throughput. Their architecture was built on TinyLlava~\citep{zhou2024tinyllava}.

\subsection{Discussion}\label{subsec5-3}
Among the various frameworks that utilize transformer architectures to address challenges in CU tasks, pre-trained architectures like Pix2Struct, TaPas and T5 show promise for CU framework development. They allow researchers to focus on novel pre-training objectives, as seen in MatCha, DePlot, and ChartT5 as they define pre-training tasks that infuse the model with essential chart comprehension skills.

Contrarily, the reliance on OCR modules within some frameworks can hinder performance due to potential errors in data extraction. In this context, researchers among the surveyed works employ an "oracle" during evaluation to isolate the effectiveness of other framework modules besides OCR. To address this limitation, frameworks like UniChart and MatCha utilize OCR-free architectures pre-trained on character-level understanding tasks, suggesting another promising direction for future research.

Another key takeaway is the inherently multi-modal nature of CU tasks, which require processing both visual and textual information. Recent advancements leverage the cross-attention technique to learn the interactions between these modalities. However, representing these modalities effectively depends on feature extraction techniques. The textual modality is often encoded using BERT-like architectures, while the visual modality utilizes more diverse techniques. Some works extract visual features using object detection and segmentation models, while others employ transformers like ViT. Finally, the tabular modality is often treated as a flattened textual sequence, in which exploration of new representations or architectural modifications for understanding data tables is needed.

A central challenge in CU lies in addressing complex reasoning tasks, which has spurred the development of prompt engineering techniques. Chain-of-thought, program-of-thought, and instruction tuning have emerged as key strategies to enhance model reasoning capabilities. While instruction tuning offers advantages in aligning models to diverse tasks, the quality of training data significantly impacts performance. The integration of foundation models has facilitated synthetic dataset generation, enabling the creation of large-scale, task-diverse training corpora.

Architectural innovations, exemplified by models like mPLUG-Owl~\citep{ye2023mplug-owl} and ChartVLM~\citep{xia2024chartx}, have introduced modularity and cascaded decoding, respectively, to improve model efficiency and reasoning abilities. Simultaneously, efforts to optimize model size and computational efficiency, as demonstrated by TinyChart, yield promising results. The interplay between model architecture, training data, and prompt engineering techniques highlights the complexity of the CU domain.

In conclusion, Table~\ref{tab5-1} summarizes the key findings of the surveyed frameworks, categorized by single-task or multi-task capabilities, across four key dimensions. The first dimension covers the supported input modality. Most frameworks handle multiple modalities as input, including chart images and textual data. However, some frameworks were tested in various settings, using either just the image or multiple modalities. For multimodal frameworks, the table modality is flattened and appended to the textual input. Notably, MatCha is the only framework that utilizes a single modality (image only) without the need for a separate branch for other modalities while handling multiple tasks. It achieves this by appending the textual information as a header to the input image. The second dimension focuses on OCR usage. Another important observation is the trend towards visual transformers, which potentially eliminates the need for OCRs due to their ability to directly process visual information. The third dimension highlights the use of newly introduced pretraining tasks specifically designed for CU modelling, either focusing on learning the underlying chart structure or equipping the model with reasoning capabilities. The fourth dimension addresses the downstream tasks tackled by each framework.

\begin{sidewaystable}
\caption{Summary of key attributes defining the transformer-based CU frameworks. Both $C_{qa}$ \& $O_{qa}$ stand for CQA and OpenCQA, respectively. $C2_x$ stands for Chart-to-$x$ where $x$ is one of the following: Text, Table, or Code. While $\phi$ stands for "other".}\label{tab5-1}
\setlength{\tabcolsep}{1pt}
\begin{tabular*}{\textwidth}{@{\extracolsep\fill}c|rcccc|cccccc}
\toprule%
\multicolumn{2}{c}{\multirow{2}{*}{Framework}} & \multirow{2}{*}{\makecell{Input \\ Modality(s)}} & \multirow{2}{*}{\makecell{OCR \\ Utilization}} & \multicolumn{2}{c}{CU Pretraining Task(s)} & \multicolumn{6}{c}{Solvable Task(s)} \\\cmidrule{5-6}\cmidrule{7-12}

\multicolumn{2}{c}{}&&& Structure & Reasoning & $C_{qa}$ & $O_{qa}$ & $C2_{txt}$ & $C2_{tab}$ & $C2_{code}$ & $\phi$ \\

\toprule
\multirow{15}{*}{\rotatebox[origin=c]{90}{Single-Task}} 
& STL-CQA~\citeyearpar{singh2020stl}          & Multimodal  & \cmark & \cmark & \xmark & \cmark & \xmark & \xmark & \xmark & \xmark & \xmark \\
& CRCT~\citeyearpar{levy2022crct}            & Multimodal  & \cmark & \xmark & \xmark & \cmark & \xmark & \xmark & \xmark & \xmark & \xmark \\
& CQAC~\citeyearpar{masry2021integrating}        & Multimodal  & \cmark & \xmark & \xmark & \cmark & \xmark & \xmark & \xmark & \xmark & \xmark \\
& TapasQA~\citeyearpar{jain2022tapasqa}         & Multimodal  & \cmark & \xmark & \xmark & \cmark & \xmark & \xmark & \xmark & \xmark & \xmark \\
& ChartQA~\citeyearpar{masry-etal-2022-chartqa}     & Multimodal  & \cmark & \xmark & \xmark & \cmark & \xmark & \xmark & \xmark & \xmark & \xmark \\
& OpenCQA~\citeyearpar{kantharaj-etal-2022-opencqa}   & Both\footnotemark[1] & \cmark & \xmark & \xmark & \xmark & \cmark & \xmark & \xmark & \xmark & \xmark \\
& Chart-to-Text~\citeyearpar{kantharaj-etal-2022-chart} & Both     & \cmark & \xmark & \xmark & \xmark & \xmark & \cmark & \xmark & \xmark & \xmark \\
& VisInsights~\citeyearpar{zhou2023intelligent}     & Single    & \xmark & \xmark & \xmark & \xmark & \xmark & \cmark & \xmark & \xmark & \xmark \\
& DePlot~\citeyearpar{liu-etal-2023-deplot}       & Single    & \xmark & \cmark & \xmark & \xmark & \xmark & \xmark & \cmark & \xmark & \xmark \\
& StructChart~\citeyearpar{xia2023structchart}     & Single    & \xmark & \cmark & \xmark & \xmark & \xmark & \xmark & \xmark & \xmark & \cmark \\
& OneChart~\citeyearpar{chen2024onechart}        & Single    & \xmark & \cmark & \xmark & \xmark & \xmark & \xmark & \xmark & \xmark & \cmark \\
& SIMPLOT~\citeyearpar{kim2024simplot}         & Single    & \xmark & \cmark & \xmark & \xmark & \xmark & \xmark & \cmark & \xmark & \xmark \\
& TTC-QuAli~\citeyearpar{dong2024ttc}          & Multimodal  & \cmark & \cmark & \xmark & \xmark & \xmark & \xmark & \xmark & \xmark & \cmark \\
& TextRoleCLS~\citeyearpar{kim2024text}         & Single    & \xmark & \xmark & \xmark & \xmark & \xmark & \xmark & \xmark & \xmark & \cmark \\
& mChartQA~\citeyearpar{wei2024mchartqa}        & Multimodal  & \xmark & \cmark & \xmark & \cmark & \xmark & \xmark & \xmark & \xmark & \xmark \\

\midrule
\multirow{12}{*}{\rotatebox[origin=c]{90}{Multi-Task}} & \multicolumn{1}{l}{\textbf{\textit{Pre-training based}}} \\
& MatCha~\citeyearpar{liu-etal-2023-matcha}       & Single    & \xmark & \cmark & \cmark & \cmark & \xmark & \cmark & \cmark & \cmark & \xmark \\
& ChartReader~\citeyearpar{cheng2023chartreader}    & Multimodal  & \xmark & \cmark & \xmark & \cmark & \xmark & \cmark & \cmark & \xmark & \xmark \\
& ChartT5~\citeyearpar{zhou2023enhanced-chartt5}    & Multimodal  & \cmark & \cmark & \xmark & \cmark & \xmark & \cmark & \xmark & \xmark & \xmark \\
\vspace{2mm}
& UniChart~\citeyearpar{masry-etal-2023-unichart}    & Multimodal  & \xmark & \cmark & \xmark & \cmark & \cmark & \cmark & \cmark & \xmark & \xmark \\
& \multicolumn{1}{l}{\textbf{\textit{Prompt-engineering based}}} \\
& ChartLlama~\citeyearpar{han2023chartllama}      & Multimodal  & \xmark & \xmark & \xmark & \cmark & \xmark & \cmark & \cmark & \cmark & \cmark \\
& MMC~\citeyearpar{liu-etal-2024-mmc}          & Multimodal  & \xmark & \cmark & \xmark & \cmark & \cmark & \xmark & \cmark & \cmark & \cmark \\
& ChartAssistant~\citeyearpar{meng2024chartassisstant} & Multimodal  & \xmark & \cmark & \xmark & \cmark & \cmark & \cmark & \cmark & \xmark & \cmark \\
& ChartInstruct~\citeyearpar{masry2024chartinstruct}  & Multimodal  & \xmark & \cmark & \xmark & \cmark & \cmark & \cmark & \xmark & \cmark & \cmark \\
& ChartVLM~\citeyearpar{xia2024chartx}         & Multimodal  & \xmark & \xmark & \xmark & \cmark & \xmark & \cmark & \xmark & \cmark & \cmark \\
& TinyChart~\citeyearpar{zhang2024tinychart}      & Multimodal  & \xmark & \xmark & \xmark & \cmark & \cmark & \cmark & \cmark & \cmark & \cmark \\

\botrule
\end{tabular*}
\footnotetext[1]{"Both" entails that different settings were attempted in their experiments. Some experiments’ settings utilized a single modality while others utilized more.}
\end{sidewaystable}

\section{Evaluation and Comparison}\label{sec6}
This section provides an overview of the performance metrics used to benchmark various aspects of a proposed framework. Previously in CU tasks such as CQA, accuracy was the primary metric used to evaluate the overall performance~\citep{kahou2017figureqa, kafle2018dvqa, singh2020stl}. However, with the emergence of solutions that handle open-vocabulary text outputs~\citep{liu-etal-2024-mmc, masry2024chartinstruct, meng2024chartassisstant}, numerical floating-point values~\citep{methani2020plotqa, masry-etal-2022-chartqa, liu-etal-2023-matcha}, text generation~\citep{kantharaj-etal-2022-chart, masry-etal-2023-unichart, han2023chartllama}, along with the table reconstruction evaluation ~\citep{liu-etal-2023-deplot, kim2024simplot}, other metrics were introduced. This section concludes with a comparative analysis of the models’ performance over well-known benchmarking sets.

\subsection{Performance Metrics}\label{subsec6-1}
\begin{enumerate}[label=\Alph*.]
  \item Chart Questions Answering
  
  \textbf{Accuracy (ACC):} Early works approaching the CQA problem as a classification task~\citep{kahou2017figureqa, kafle2018dvqa, singh2020stl} compute the number of exact matches between predicted and ground truth answer. This method does not account for errors stemming from OCR or close numerical approximations.

  \textbf{Relaxed Accuracy (RA):} Used to address the limitations of exact matching, allowing for a small marginal error of 5\%~\citep{methani2020plotqa}. RA has subsequently become the standard metric for evaluating numerical answers~\citep{masry-etal-2022-chartqa, cheng2023chartreader, zhou2023enhanced-chartt5, wei2024mchartqa}. 

  For textual output, either accuracy is used, or other metrics such as edit distance can be adopted.
 
  \vspace{5mm}
  \item Table Reconstruction

  \textbf{Relative Number Set Similarity (RNSS):}~\cite{masry-etal-2022-chartqa} extend the metrics of~\citep{luo2021chartocr} for evaluating how close the extracted data is to the ground truth. The metric compares the predicted data values with the ground truth, independent of their organization in the data table.  This does not account for the overall structure of the data in a table.
  
  \textbf{Relative Mapping Similarity (RMS):} To address the flaws of RNSS, \cite{liu-etal-2023-deplot} introduced RMS. This metric computes the similarity of the whole extracted data table while being insensitive to the order of the columns and rows. It addresses extracted textual data, and tolerates minor errors in the extracted numeric or textual data within a margin. Additionally, Normalized Levenshtein Distance ($NL_\tau$)~\citep{biten2019scene-leven-dist} is measured for textual data entries to evaluate their dissimilarity and allow minor character differences. 
  
  \textbf{Structuring Chart-oriented Representation Metric (SCRM):}~\cite{xia2023structchart} proposed SCRM, a novel metric designed to evaluate the effectiveness of chart derendering methods. SCRM assesses the degree to which a generated representation captures the structural information present in the original chart. The STR tuple $1^{st}$ and $2^{nd}$ entities (i.e., row and column headers, respectively) are treated as strings, while the value floats. It offers an image-wise STR similarity level metric (i.e. Intersection over Union (IoU)), and two dataset-wise metrics to summarize the IoU across all images (i.e. Precision and mPrecision). 

  \textbf{Relative Distance (RD):} The RMS metric, a product of Levenshtein distance and relative distance, tends to underperform in the chart-to-table task when textual differences are produced. To address this, \citep{kim2024simplot} proposed RD. This metric is derived by constructing a Levenshtein distance matrix from concatenated column and row predictions and ground truths, followed by minimal cost matching. By focusing solely on relative distance, RD provides a more accurate evaluation of value mapping performance.
  
  \vspace{5mm}
  \item Natural Language Generation
  
  \textbf{ROUGE:} Introduced by~\citep{lin2004rouge}, ROUGE is a set of recall-based metrics for assessing the quality of the generated text. Unlike accuracy metrics, which focus on a single correct answer, ROUGE emphasizes the similarity between the model's output and human-written reference summaries. It achieves this by calculating the n-gram overlap between the generated text and the reference summaries. The resulting score is expressed as a percentage, indicating how many n-grams from the reference summaries are also present in the model's output.

  \textbf{CIDEr:} In order to assess the quality of generated image captions, \citep{vedantam2015cider} introduced the CIDEr metric to measure the similarity between a generated text description and a set of human-written reference captions. It achieves this by calculating a score based on the weighted overlap of n-grams between the generated text and the reference captions. Here, the weights are determined using TF-IDF, which assigns higher importance to n-grams that are frequent in a single caption but rare across the entire reference set.

  \textbf{BLEU:} Another n-gram matching metric commonly used is BLEU~\citep{papineni2002bleu}. However, as~\cite{post2018call-bleu4} argues, BLEU should not be treated as a single, definitive metric. Authors should clearly specify the parameters they use when evaluating with BLEU. While a high BLEU score indicates a high degree of n-gram overlap between the generated text and the reference, it does not necessarily guarantee semantic equivalence or factual accuracy. The metric is primarily precision-oriented, meaning it may favour summaries that include all the information from the reference text, even if phrased differently. This can potentially overlook fluency or factual correctness.

  Traditional n-gram-based metrics, which focus on counting matching phrases, have long been used for text generation evaluation. However, a more nuanced approach has emerged to better reflect human perception of quality. Such metrics are BERTScore~\citep{zhang2019bertscore} and BLEURT~\citep{sellam-etal-2020-bleurt}.

  \textbf{Content Selection (CS):} One of the comparison strategies introduced by~\citep{wiseman-etal-2017-challenges-cs} is CS. It evaluates the quality of extractive summaries in comparison with a reference summary. CS depends on an earlier stage of information extraction in which entities and their relations are introduced. It then measures how well the generated summary aligns with the ground-truth summary by calculating the precision and recall of unique relations extracted from both summaries.

  \textbf{BERTScore:} Introduced by~\citep{zhang2019bertscore}, BERTScore offers a more nuanced approach by calculating token-level similarity scores between the generated and reference texts. Unlike metrics that rely solely on exact matches, BERTScore leverages contextual embeddings to capture semantic equivalence even when phrased differently. This allows for a stronger correlation with human judgments of quality.

  \textbf{BLEURT:} Similar to BERTScore, BLEURT \citep{sellam-etal-2020-bleurt}, is another metric that measures semantic equivalence. It also leverages pre-trained BERT embeddings and computes the similarity score of sentences as the sum of cosine similarities between their tokens' embeddings.

\end{enumerate}

\subsection{Comparative Analysis}\label{subsec6-2}
This section presents a comparative analysis of the reviewed CU frameworks across core tasks: CQA, chart-to-text generation, and chart derendering. CQA performance is evaluated on FigureQA, DVQA, ChartQA, and PlotQA datasets using Relaxed Accuracy (RA) for factoid questions, and on the OpenCQA dataset using BLEU-4 \citep{post2018call-bleu4} for open-ended questions. While a dedicated benchmark for chart derendering tasks is absent, the chart-to-table task, primarily evaluated on ChartQA using RNSS~\citep{masry-etal-2022-chartqa}, RMS~\citep{liu-etal-2023-deplot}, and RD~\citep{kim2024simplot} metrics, serves as a proxy. Chart-to-text models are assessed on subsets (i.e. Statista and Pew) of the Chart-to-Text~\citep{kantharaj-etal-2022-chart} corpus (detailed in Section~\ref{subsec4-7}) using BLEU-4 to measure the quality of generated text summaries.

\vspace{5mm}
\noindent\textbf{Chart-to-Table.} Both Deplot~\citep{liu-etal-2023-deplot} and SIMPLOT~\citep{kim2024simplot} function as specialized chart-to-table modules, compatible with various LLMs or MLLMs. These frameworks generate data tables, leaving subsequent reasoning processes to downstream components. In contrast, frameworks such as MatCha~\citep{liu-etal-2023-matcha}, UniChart~\citep{masry-etal-2023-unichart}, ChartAssistant \citep{meng2024chartassisstant}, and TinyChart~\citep{zhang2024tinychart} adopt an E2E approach, integrating chart-to-table conversion within a comprehensive model architecture. The comparative performance of these frameworks on the aforementioned benchmark and metrics is summarized in Table~\ref{tab6-1}. Other frameworks that were concerned with the chart-to-table task are ChartT5~\citep{zhou2023enhanced-chartt5} and ChartReader~\citep{cheng2023chartreader}. However, ChartT5 only observed the effects of the pre-training objectives on the downstream task, without reporting the quality of the retrieved tables in comparison to the ground-truth ones. Unlike ChartT5, ChartReader reported the quality of their extracted tables, however, it was not benchmarked using the aforementioned metrics, but contrarily on the earlier version of RNSS introduced by~\citep{luo2021chartocr}.

\begin{table}[h]
\captionsetup{width=\textwidth}
\caption{Chart-to-Table performance evaluation on the ChartQA dataset. The reported results are using the RNSS, RMS$_\text{F1}$ and RD$_\text{F1}$ metrics. Bold and underlined values indicate top and second-best performance, respectively. For frameworks offering multiple variants, the highest performing variant for the specified task is presented.}\label{tab6-1}%
\begin{tabular}{@{}l|ccc@{}}
\toprule
 & \multicolumn{3}{c}{ChartQA}\\\cmidrule{2-4}
 Method $\downarrow$ & RNSS $\uparrow$ & RMS$_\text{F1}$ $\uparrow$ & RD$_\text{F1}$ $\uparrow$ \\
\midrule
MatCha~\citeyearpar{liu-etal-2023-matcha} & 85.21  & 83.49 & \textbf{--}  \\
DePlot~\citeyearpar{liu-etal-2023-deplot} & \textbf{97.10}  & \textbf{94.20} & \underline{90.95} \\
UniChart~\citeyearpar{masry-etal-2023-unichart} & \underline{94.01}  & 91.10 & 88.03 \\
ChartAst$_\text{D}$~\citeyearpar{meng2024chartassisstant} & \textbf{--}  & 92.00 & \textbf{--}  \\
TinyChart~\citeyearpar{zhang2024tinychart} & \textbf{--}  & \underline{93.78} & \textbf{--} \\
SIMPLOT~\citeyearpar{kim2024simplot}    & \textbf{--}  & \textbf{--} & \textbf{92.32} \\
\botrule
\end{tabular}
\end{table}

\vspace{5mm}
\noindent\textbf{Chart Question Answering.} Among the single-task frameworks dedicated to factoid chart question answering, including STL-CQA~\citep{singh2020stl}, CRCT~\citep{levy2022crct}, ChartQA~\citep{masry-etal-2022-chartqa}, and mChartQA~\citep{wei2024mchartqa}, the latter demonstrates superior performance. This enhancement could be attributed to mChartQA's training strategy, specifically designed to address complex reasoning questions necessitating multi-step inferencing. Among the multi-task frameworks, TinyChart~\citep{zhang2024tinychart} exhibits superior performance, surpassing a 70\% accuracy threshold on the challenging human-authored ChartQA dataset. This advancement could be attributed to the model's innovative program-of-thought training strategy. The comparative performance of these frameworks (in addition to the ones previously covered in Section~\ref{subsec5-2}) on the aforementioned benchmarks and metrics is summarized in Table~\ref{tab6-2}. 

In contrast to the end-to-end models, the models of DePlot~\citep{liu-etal-2023-deplot}, StructChart~\citep{xia2023structchart}, and SIMPLOT~\citep{kim2024simplot}, demonstrated comparable performance especially when integrated with LLMs such as GPT-3.5~\citep{ouyang2022training-gpt3-5} and when complemented by prompt engineering techniques like chain-of-thought. Through this approach, SIMPLOT attained an accuracy of 78.07\% on the human-authored challenge, 88.42\% on the augmented dataset, and an overall accuracy of 83.24\%.

Concerning the open-ended chart question answering, the OpenCQA challenge lacked dedicated frameworks. Among the reviewed models, TinyChart~\citep{zhang2024tinychart} achieved a BLEU-4 score of 20.39, likely attributed to its program-of-thought training approach. Prior to TinyChart, ChartInstruct~\citep{masry2024chartinstruct} secured the second-best performance with a BLEU-4 score of 16.71, potentially due to its instruction-tuning strategy. The comparative performance of these frameworks (in addition to the ones previously covered in Section~\ref{subsec5-2}) on the aforementioned benchmark and metrics is summarized in Table~\ref{tab6-3}.

\begin{table}[h]
\caption{CQA “factoid” performance evaluation using Relaxed Accuracy (RA). Results for recent models on FigureQA, DVQA, PlotQA, and ChartQA datasets are presented. FigureQA includes average results for both validation (V avg) and test (T avg) sets. DVQA results cover test-familiar (TF) and test-novel (TN) subsets. PlotQA offers two validation sets (V1, V2). ChartQA provides both machine-augmented (aug) and human-authored (hum) test sets (“avg” represents the average). Bold and underlined values indicate top and second-best performance, respectively. For frameworks offering multiple variants, the highest performing variant for the specified task is presented.}\label{tab6-2}
\setlength{\tabcolsep}{1pt}
\begin{tabular*}{\textwidth}{@{\extracolsep\fill}l|ccccccccc}
\toprule%
& \multicolumn{2}{@{}c@{}}{FigureQA} & \multicolumn{2}{@{}c@{}}{DVQA} & \multicolumn{2}{@{}c@{}}{PlotQA} & \multicolumn{3}{@{}c@{}}{ChartQA} 
\\\cmidrule{2-3}\cmidrule{4-5}\cmidrule{6-7}\cmidrule{8-10}%
Method $\downarrow$ & V avg & T avg & TF & TN & V1 & V2 & aug & hum & avg \\
\midrule
Human Baseline $\rightarrow$ & -- & -- & -- & 88.18 & -- & -- & -- & -- & -- \\
\midrule
STL-CQA~\citeyearpar{singh2020stl}               & -- & -- & \textbf{97.35} & \textbf{97.51} & -- & -- & -- & --& -- \\
CRCT~\citeyearpar{levy2022crct}                 & 89.83 & 89.50 & -- & -- & 76.90 & 34.40 & -- & -- & -- \\ 
VisionTaPas-OCR~\citeyearpar{masry-etal-2022-chartqa}      & 91.46 & 90.66 & 94.43 & 94.54 & 65.30 & 42.50 & 29.60 & 61.44 & 45.42 \\
Pix2Struct-\textit{base}~\citeyearpar{lee2023pix2struct}    & -- & -- & -- & -- & 73.20 & 71.90 & 81.60 & 30.50 & 56.05 \\
MatCha~\citeyearpar{liu-etal-2023-matcha}            & -- & -- & -- & -- & \textbf{92.30} & \textbf{90.70} & 90.20 & 38.20 & 64.20 \\
ChartReader-T5~\citeyearpar{cheng2023chartreader}        & \underline{95.65} & \underline{93.90} & 95.40 & 96.50 & 78.10 & 59.30 & -- & -- & 52.60 \\
ChartT5~\citeyearpar{zhou2023enhanced-chartt5}         & -- & -- & -- & -- & -- & -- & 74.40 & 31.80 & 53.10 \\
UniChart~\citeyearpar{masry-etal-2023-unichart}         & -- & -- & -- & -- & -- & -- & 88.56 & 43.92 & 66.24 \\
ChartLlama~\citeyearpar{han2023chartllama}           & -- & -- & -- & -- & -- & -- & 90.36 & 48.96 & 69.66 \\
MMCA~\citeyearpar{liu-etal-2024-mmc}              & -- & -- & -- & -- & -- & -- & -- & -- & 57.40 \\
ChartAst$_\text{S}$~\citeyearpar{meng2024chartassisstant}    & -- & -- & -- & -- & -- & -- & \textbf{93.90} & 65.90 & \underline{79.90} \\
mChartQA$_\text{Intern-LM2}$~\citeyearpar{wei2024mchartqa}   & \textbf{96.06} & \textbf{96.30} & -- & -- & \underline{78.25} & \underline{74.79} & 89.76 & \underline{68.24} & 79.00 \\
ChartInstruct~\citeyearpar{masry2024chartinstruct}       & -- & -- & -- & -- & -- & -- & 93.84 & 50.16 & 72.00 \\
TinyChart~\citeyearpar{zhang2024tinychart}           & -- & -- & -- & -- & -- & -- & \underline{93.86} & \textbf{73.34} & \textbf{83.60} \\
\botrule
\end{tabular*}
\end{table}

\begin{table}[h]
\captionsetup{width=\textwidth}
\caption{CQA “open-ended” performance evaluation on OpenCQA using BLEU4 metric. These models are of a generative nature. Bold and underlined values indicate top and second-best performance, respectively. For frameworks offering multiple variants, the highest performing variant for the specified task is presented.}\label{tab6-3}%
\begin{tabular}{@{}l|c@{}}
\toprule
Method $\downarrow$ & OpenCQA\\
\midrule
T5~\citeyearpar{masry-etal-2022-chartqa} & 9.28 \\
VL-T5~\citeyearpar{masry-etal-2022-chartqa} & 14.73 \\
UniChart~\citeyearpar{masry-etal-2023-unichart} & 14.88 \\
ChartAst$_\text{S}$~\citeyearpar{meng2024chartassisstant} & 15.50 \\
ChartInstruct~\citeyearpar{masry2024chartinstruct} & \underline{16.71} \\
TinyChart~\citeyearpar{zhang2024tinychart}     & \textbf{20.39}  \\
\botrule
\end{tabular}
\end{table}

\vspace{5mm}
\noindent\textbf{Chart-to-Text.} Regarding the Pew challenge set, TinyChart~\citep{zhang2024tinychart} demonstrated SoTA performance, attaining a BLEU-4 score of 17.93, surpassing previous models. Conversely, on the Statista challenge set, ChartReader~\citep{cheng2023chartreader} achieved a BLEU-4 score of 44.20, slightly outperforming ChartInstruct~\citep{masry2024chartinstruct}, which attained a score of 43.53. The comparative performance of these frameworks (in addition to the ones previously covered in Section~\ref{subsec5-2}) on the aforementioned benchmark and metrics is summarized in Table~\ref{tab6-4}.

\begin{table}[h]
\captionsetup{width=\textwidth}
\caption{Chart-to-Text task using BLEU4 metric. Frameworks with an “$\clubsuit$” do not explicitly specify which BLEU parameterization was used. Bold and underlined values indicate top and second-best performance, respectively. For frameworks offering multiple variants, the highest performing variant for the specified task is presented.}\label{tab6-4}%
\begin{tabular}{@{}l|cc@{}}
\toprule
 & \multicolumn{2}{c}{Chart-to-Text}\\\cmidrule{2-3}
 Method $\downarrow$ & Statista & Pew \\
\midrule
OCR-T5$^\clubsuit$~\citeyearpar{kantharaj-etal-2022-chart} & 35.29 & 10.49 \\
MatCha~\citeyearpar{liu-etal-2023-matcha} & 39.40 & 12.20 \\
ChartReader~\citeyearpar{cheng2023chartreader} & \textbf{44.20} & 14.20 \\
ChartT5~\citeyearpar{zhou2023enhanced-chartt5} & 37.51 & 09.05 \\
UniChart$^\clubsuit$~\citeyearpar{masry-etal-2023-unichart} & 38.21  & 12.48 \\
ChartLlama~\citeyearpar{han2023chartllama} & 40.71 & 14.23 \\
ChartAst$_\text{S}$~\citeyearpar{meng2024chartassisstant} & 41.00  & \underline{15.20}  \\
ChartInstruct~\citeyearpar{masry2024chartinstruct} & \underline{43.53}  & 13.83 \\
TinyChart~\citeyearpar{zhang2024tinychart}     & \textbf{--}  & \textbf{17.93} \\
\botrule
\end{tabular}
\end{table}

\subsection{Discussion}\label{subsec6-3}
Attentive evaluation is essential for assessing the efficacy of novel CU frameworks. While the development of dedicated benchmarks is valuable, benchmarking against existing standards provides a crucial reference point. Our analysis reveals that the CQA task, particularly for human-authored queries, remains challenging, with SIMPLOT achieving a SoTA accuracy of 78.07\%. Frameworks incorporating chart derendering among prompt-engineering techniques, exemplified by TinyChart, demonstrate superior performance across multiple CU tasks, suggesting a strong correlation between accurate table reconstruction and CQA success. While LLMs show promise in chart-to-text and open-ended CQA, the over-reliance on metrics like BLEU may be insufficient. A comprehensive evaluation necessitates diverse metrics to assess factual accuracy and fluency. Establishing robust benchmarks and metrics is paramount for advancing the field and enabling meaningful comparisons between frameworks.


\section{Applications, Challenges, and Future Directions}\label{sec7}
The applications of CU encompass a wide range of domains. This section explores four key areas: chart fact-checking, domain-specific agents, chart modification and accessibility. To fully realize the potential of CU, we identify nine open challenges and future directions. These include issues related to visual and numerical reasoning, image resolution, model interpretability, and deployment considerations. Addressing these obstacles will pave the way for more robust and impactful chart understanding systems.

\subsection{Applications}\label{subsec7-1}
\subsubsection{Chart Fact-Checking}\label{subsubsec7-1-1}
Misinformation and disinformation pose significant challenges in contemporary society. The proliferation of social media has amplified the dissemination of such content. Financial, social, and political motivations promote the creation of online misinformation~\citep{CHAUDHURI2024102654misinfo}. \cite{akhtar2023reading} highlight the role of chart images in disseminating misinformation. To address this issue, they introduce the ChartFC dataset, which enables the classification of chart images as supporting or refuting textual claims. Subsequently, \cite{akhtar2024chartcheck} introduce a more complex dataset requiring models to provide explanations for their classifications. Each claim in this dataset is accompanied by a contextual explanation. 

Misinformation can also originate from chart captions. To address this, \citep{huang2023lvlms} introduce a chart caption factual error correction task. This task aims to identify and correct inaccuracies in generated captions. They also introduce the ChartVE metric to evaluate caption consistency with the chart image. Similarly, \cite{krichene2024faithful} propose the ChaTS-Critic metric. This metric derenders the chart image into a data table and analyzes the caption sentence-by-sentence to identify inconsistencies. This research area holds promise for developing tools to combat misinformation spread through social media.

\subsubsection{Domain-Specific Agents}\label{subsubsec7-1-2}
Charts serve as fundamental tools for visualizing and interpreting data across various domains. In finance, charts support portfolio analysis, performance evaluation, and risk assessment~\citep{dao2024ai}. Within the medical field, charts aid in understanding complex medical data, including vital signs, lab results, and epidemiological data~\citep{waqas2024control}. Charts also enhance learning in education by visualizing data across mathematics, language arts, social sciences, and natural sciences. By transforming numerical data into visual patterns, charts facilitate informed decision-making, knowledge acquisition, and comprehension in these respective domains.

Domain-specific agents capable of comprehending chart images hold significant potential across various sectors. In finance, these agents could facilitate investment analysis, risk assessment, and portfolio management. Within healthcare, they can contribute to medical diagnosis, treatment planning, and epidemiological research. In education, such agents could revolutionize learning by providing interactive explanations, answering student queries, and personalizing educational content~\citep{kim2024chatgpt}. These applications underscore the need for further development of robust chart understanding models to unlock the full potential of visual data in these domains.

\subsubsection{Chart Modification}\label{subsubsec7-1-3}
Chart modification encompasses image-to-image tasks, primarily involving chart editing or reconstruction. ChartLlama~\citep{han2023chartllama} introduces a chart-editing task requiring a chart image and an editing instruction as input. This could be used to evaluate a model’s ability to generate a modified chart image aligned with the provided instruction. Similarly, Chart X~\citep{xia2024chartx} introduces chart redrawing, focusing on replicating chart structure while preserving the original chart type based on the provided data. These tasks collectively contribute to advancing chart understanding and manipulation capabilities. The integration of these tasks within chart recommendation or generation systems has the potential to streamline the design process for end-users.

\subsubsection{Accessible Chart Interfaces}\label{subsubsec7-1-4}
Approximately 2.2 billion individuals worldwide experience visual impairments, according to the World Health Organization~\citep{WHO2023}. Screen readers serve as indispensable tools for Blind and Visually Impaired (BVI) users navigating digital environments~\citep{shahira2021towards}. Integrating chart understanding tasks, including derendering, question answering, and text generation, within assistive technologies can significantly enhance accessibility for BVI users. On that end, \cite{kim2023exploring} conducted a case study examining design considerations for developing question-answering systems tailored to BVI users. This research lays the groundwork for future accessible chart question-answering systems. Regarding the impact of chart derendering on accessibility, the SVG format offers advantages for screen readers and conversion to tactile formats for visually impaired users. Chart4Blind~\citep{moured2024chart4blind} evaluates a prototype system for converting bitmap charts into accessible SVG images, demonstrating improved efficiency and effectiveness for both sighted and visually impaired users.

More recent works on automatic caption generation systems include case studies assessing user needs, expectations, and system utility~\citep{choi2022intentable, hsu2024scicapenter, singh2024figura11y}. \cite{choi2022intentable} developed Intentable; an intent-based system for caption generation, enabling users to interactively guide the captioning process according to desired levels of chart insight coverage. A case study conducted by~\citep{hsu2024scicapenter} investigates the applicability of automated chart captioning within scholarly articles through their SciCapenter system. The research evaluates the utility of machine-generated caption ratings relative to generated captions, employing an analysis checklist. Findings indicate a preference for automated caption ratings over generation, suggesting potential for developing metrics to assess caption effectiveness. FigureA11y~\citep{singh2024figura11y}, another end-to-end alt text generation system, was investigated to compare user preferences between fully automated and interactive caption generation. The study revealed a preference for interactive systems offering additional features like cursor-based generation and user query capabilities. Language accessibility represents another critical dimension. While most research focuses on English, OneChart~\citep{chen2024onechart} is the first dataset to offer both English and Chinese language support.

\subsection{Challenges \& Future Directions}\label{subsec7-2}
Despite the remarkable progress achieved by transformer-based models in CU, several challenges persist. These limitations underscore the need for continued research to enhance model capabilities and facilitate real-world applications. The following paragraphs delve into the primary challenges and explore promising avenues for future development within the CU domain.

\textbf{OCR Dependency.} A prevalent challenge among reviewed frameworks is the reliance on external OCR modules, which often exhibit suboptimal performance due to the distinctive layout characteristics of chart images. To address this limitation, recent studies, including DePlot~\citep{liu-etal-2023-deplot}, ChartAssistant~\citep{meng2024chartassisstant}, and ChartInstruct~\citep{masry2024chartinstruct}, have explored OCR-free approaches. The integration of VLPs and MLLMs further demonstrates the potential for OCR-independent chart understanding.

\textbf{Input Image Resolution.} Existing research predominantly employs high-resolution chart images, neglecting the challenges posed by low-resolution scanned counterparts prevalent in real-world applications~\citep{masry-etal-2023-unichart, zhang2024tinychart}. The development of a benchmark dataset comprising scanned visualizations is imperative. Alternatively, advancements in super-resolution techniques offer a potential solution.

\textbf{Visual Reasoning.} Certain frameworks, including DePlot~\citep{liu-etal-2023-deplot}, exhibit limited utilization of visual features within their reasoning components. While DePlot exclusively relies on LLMs for processing reconstructed data tables, ChartReader~\citep{cheng2023chartreader}, ChartT5~\citep{zhou2023enhanced-chartt5}, and SIMPLOT~\citep{kim2024simplot} incorporate cross-modal techniques, demonstrating the efficacy of fusing visual and textual information. Continued research on multimodal data fusion is imperative for advancing model performance.

\textbf{Hyper-parameter Tuning.} The optimization of hyperparameters is often constrained by computational resource limitations and prolonged training times, particularly for large-scale models. While training strategies were not explicitly detailed in the reviewed studies, the adoption of distributed training and sparse data representations could potentially accelerate the hyperparameter tuning process and facilitate comprehensive exploration of the hyperparameter space.

\textbf{Densely Populated Charts.} Existing research encounters limitations when processing charts with high information density. UniChart~\citep{masry-etal-2023-unichart} exemplifies this challenge, demonstrating reduced prediction accuracy in such scenarios. Overcoming these limitations necessitates further research to enhance model performance in handling complex chart visualizations.

\textbf{Numerical Reasoning.} As detailed in Section~\ref{subsec6-3}, human-authored queries continue to pose significant challenges for CQA systems. Despite advancements, current SoTA models, such as SIMPLOT, achieve accuracy rates of only 78.07\%, indicating ample room for improvement. The two-stage training paradigm of SIMPLOT, which distills knowledge from a chart-to-table model, contributes to its relatively strong performance. Prompt engineering techniques, as exemplified by ChartInstruct~\citep{masry2024chartinstruct} and TinyChart~\citep{zhang2024tinychart}, have also yielded substantial advancements. To further enhance CQA capabilities, future research should explore the integration of mathematical and logical reasoning challenges into model training. This approach could strengthen the model's capacity to address complex numerical reasoning inherent in human-authored queries.

\textbf{Model Efficiency.} Large transformer models, demonstrated by Pix2Struct~\citep{lee2023pix2struct} with its substantial 5GB footprint, present significant computational challenges. While UniChart~\citep{masry-etal-2023-unichart} achieves a 28\% reduction to 201 million parameters, further optimization is necessary. Knowledge distillation~\citep{hinton2015distilling} offers a promising avenue for minimizing model size and accelerating inference. The successful application of SIMPLOT reduced parameters from 1.3 billion to 374 million, underscores its potential~\citep{kim2024simplot}.

\textbf{Model Collapse.} While recent frameworks such as ChartAst, ChartLlama, and ChartInstruct leverage foundation models for synthetic data generation, the potential for model collapse due to the indiscriminate use of model-generated content remains a concern~\citep{shumailov2024ai}. The erosion of data diversity and the emergence of distributional shifts pose significant challenges to the long-term efficacy of these approaches.

\textbf{Explainable AI (XAI).} Recent studies~\citep{zhou2023intelligent, wang2024salchartqa} have employed saliency maps to enhance chart captioning and question answering performance by identifying relevant chart regions. These findings underscore the significance of information-seeking behaviors in shaping attention mechanisms and open avenues for task-driven visualization optimization and explainable AI within the domain of chart understanding.

\section{Conclusion}\label{sec8}
This review article explored the advancements and current challenges in CU research. We examined the utilization of transformer models and various frameworks designed for CU tasks. An overview of leading architectures was provided, including ViT, VL-T5, VisionTapas, Pix2Struct, and their variations, highlighting their strengths and limitations. Additionally, we highlighted the current benchmarking datasets, including the newly introduced OpenCQA dataset, and discussed the challenges they present. The analysis of these datasets revealed the importance of data diversity, well-defined formats, and element annotations for facilitating different CU tasks. We also analyzed the different pre-training tasks employed to inject chart comprehension and mathematical reasoning capabilities into the models.

Despite significant advancements in CU driven by transformer-based models, several key challenges persist. OCR dependency, limited handling of low-resolution images, and the need for enhanced visual reasoning capabilities remain focal points. Moreover, the development of robust benchmarks and the optimization of model efficiency are essential for advancing the field. The integration of explainable AI techniques is crucial for understanding model decision-making processes and fostering trust in CU systems. Addressing these challenges will pave the way for the development of more sophisticated and versatile chart understanding models with broader real-world applications.

While progress has been made in generating synthetic data using foundation models, concerns regarding model collapse and the erosion of data diversity warrant further investigation. A balanced approach that combines synthetic and real-world data is likely necessary to mitigate these risks. Overall, the CU domain is a dynamic and evolving field with substantial potential for future research and innovation.

\section*{Declarations}

\subsection*{Ethics Statement}\label{subsecA-2}
To ensure academic integrity, all figures comply with relevant copyright regulations. Furthermore, the numerical values in Tables \ref{tab6-1}, \ref{tab6-2}, \ref{tab6-3}, and \ref{tab6-4} are derived from a comprehensive analysis of the included studies. Google Gemini is used in proofreading the manuscript.

\bibliography{sn-bibliography}

\end{document}